\RequirePackage{fix-cm}
\documentclass[twocolumn]{svjour3}          
\smartqed  
\usepackage{graphicx}

\usepackage{pgfplots}
\usepackage{pgfplots}  

\usepackage{url}
\usepackage{array}
\usepackage{subfigure}
\usepackage{amsmath}
\usepackage{float}
\usepackage{color}
\usepackage{amsmath}

\usepackage{booktabs}
\usepackage{multirow} 
\usepackage{ulem}

\usepackage[breaklinks=true,colorlinks,citecolor=blue,urlcolor=blue,linkcolor=blue,bookmarks=false,pagebackref=true]{hyperref}
\usepackage{amssymb}

\newcolumntype{M}[1]{>{\centering\arraybackslash}m{#1}}

\newcommand{\etal}{\textit{et al}.}
\newcommand{\ie}{\textit{i}.\textit{e}.}
\newcommand{\eg}{\textit{e}.\textit{g}.}

\begin{document}

\title{Deep Image Deblurring: A Survey
}


\author{Kaihao Zhang         \and
        Wenqi Ren        \and
        Wenhan Luo        \and
        Wei-Sheng Lai  \and \\
        Bj\"orn Stenger  \and
        Ming-Hsuan Yang   \and
        Hongdong Li
}

\institute{Kaihao Zhang and Hongdong Li \at
        Australian National University, Australia\\
        \email{\{kaihao.zhang, hongdong.li\}@anu.edu.au}        \\
        \and
        Wenqi Ren and Wenhan Luo\at
        Sun Yat-sen University, Guangzhou, 510275, China\\
        \email{\{rwq.renwenqi, whluo.china\}@gmail.com} \\
        \and
         Bj\"orn Stenger \at
         Rakuten Institute of Technology, Rakuten Group Inc. \\
         \email{bjorn@cantab.net}        \\
         \and
         Weisheng Lai and Ming-Hsuan Yang \at
         School of Engineering, University of California at Merced, Merced, CA, USA \\
           \email{\{wlai24, mhyang\}@ucmerced.edu} \\
}

\date{Received: date / Accepted: date}

\maketitle

\begin{abstract}
Image deblurring is a classic problem in low-level computer vision with the aim to recover a sharp image from a blurred input image.  Advances in deep learning have led to significant progress in solving this problem, and a large number of deblurring networks have been proposed.  This paper presents a comprehensive and timely survey of recently published deep-learning based image deblurring approaches, aiming to serve the community as a useful literature review.  
We start by discussing common causes of image blur, 
introduce benchmark datasets and performance metrics, and summarize different problem formulations.
Next, we present a taxonomy of methods using convolutional neural networks (CNN) based on architecture, loss function, and application, offering a detailed review and comparison.
In addition, we discuss some domain-specific deblurring applications including face images, text, and stereo image pairs.  We conclude by discussing key challenges and future research directions. 
\end{abstract}

\section{INTRODUCTION}
\begin{figure}[!tb]
  \centering
  \subfigure[Camera shake blur]{
    \label{idea:a}
    \includegraphics[width=0.48\linewidth ]{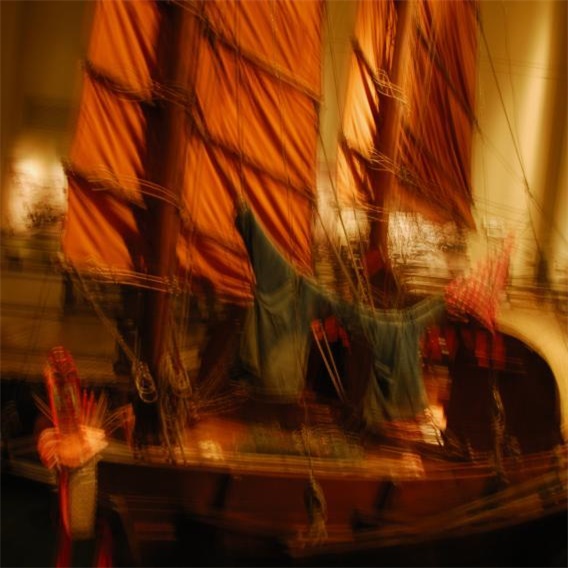}}
  \subfigure[Out-of-focus blur]{
    \label{idea:b}
    \includegraphics[width=0.48\linewidth ]{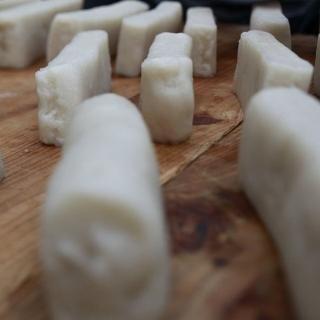}}
    \subfigure[Moving object blur]{
    \label{idea:c}
    \includegraphics[width=0.48\linewidth ]{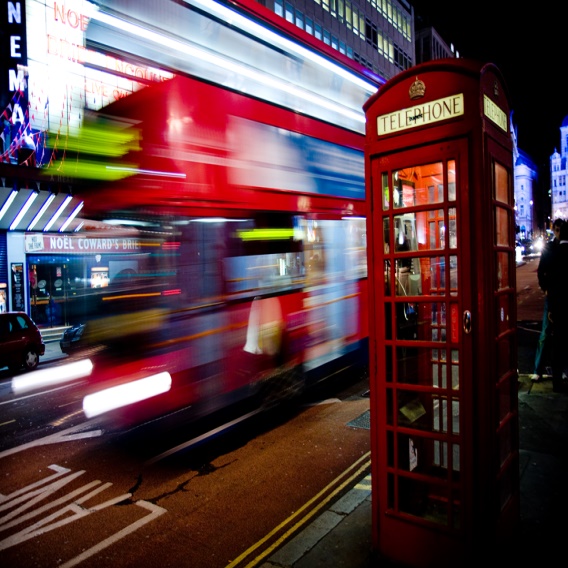}}
  \subfigure[Mixed blur]{
    \label{idea:d}
    \includegraphics[width=0.48\linewidth ]{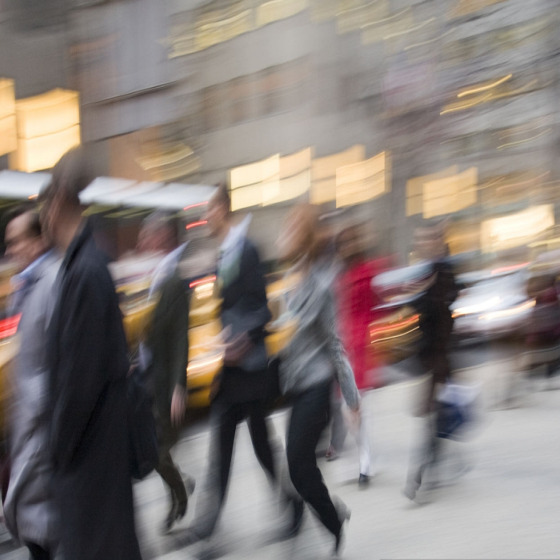}}
\caption{Examples of different blurry images. Causes for blur include (a) camera shake, (b) out-of-focus scene, (c) moving objects, and (d) multiple causes, respectively.}
  \label{different_blur}
\end{figure}

Image deblurring is a classic task in low-level computer vision,  which has attracted the attention from the image processing and computer vision community. The objective of image deblurring is to recover a sharp image from a blurred input image, where the blur can be caused by various factors such as lack of focus, camera shake, or fast target motion \cite{Abuolaim2020defocus,chen2015multispectral,kang2007automatic,sun2015learning}. Some examples are given in Figure~\ref{different_blur}.

\begin{figure*}[!tb]
  \centering
    \includegraphics[width=0.99\linewidth ]{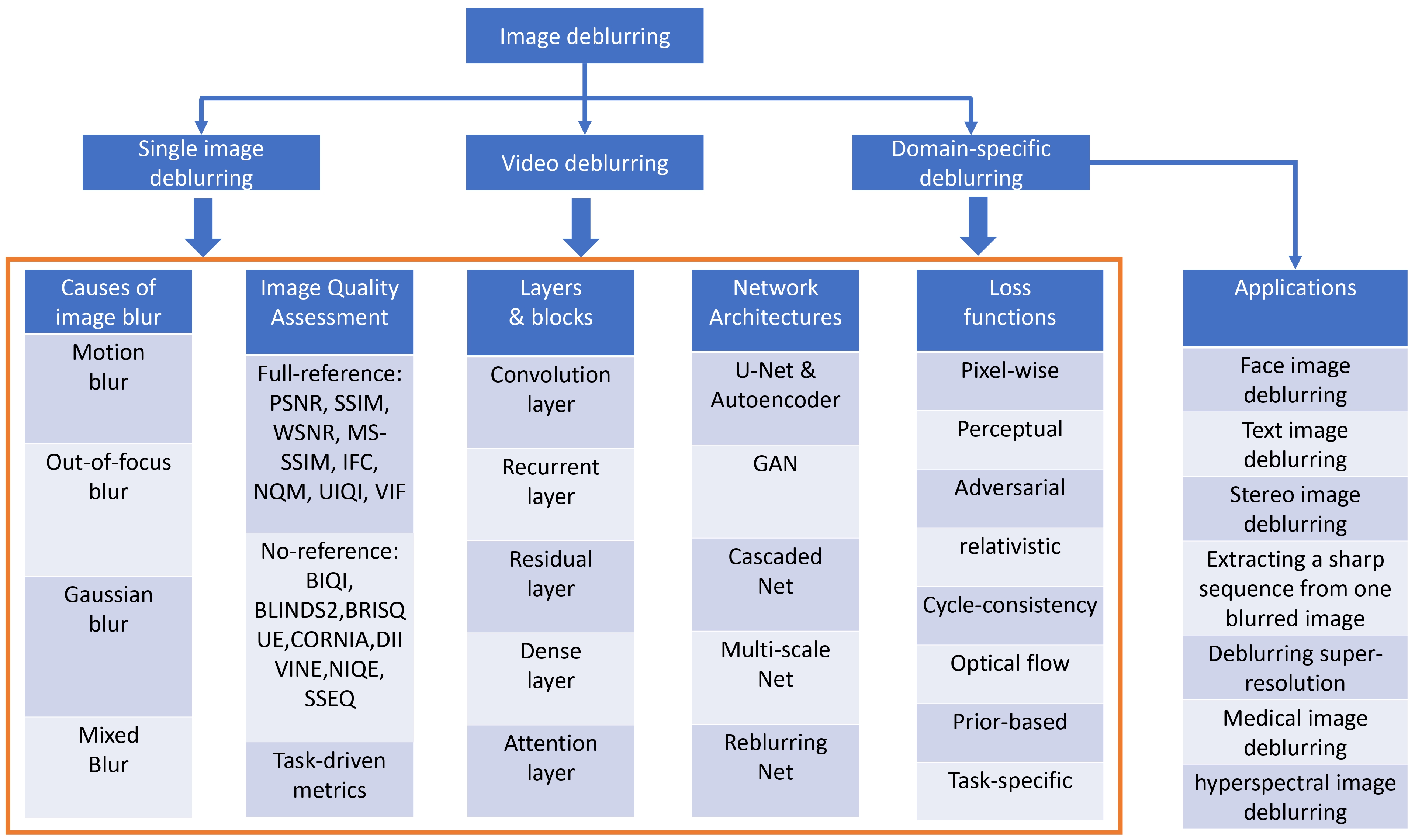}
\caption{Taxonomy of existing deep image deblurring techniques reviewed in this survey.}
  \label{taxonomy}
\end{figure*}

Non-deep learning image deblurring methods often formulate the task as an inverse filtering problem, where a blurred image is modeled as the result of the convolution with blur kernels, either spatially invariant or spatially varying. Some early approaches assume that the blur kernel is known, and adopt classical image deconvolution algorithms such as Lucy-Richardson, or Wiener deconvolution, with or without Tikhonov regularization, to restore sharp images \cite{schmidt2013discriminative,szeliski2010computer,xu2014inverse}. 
On the other hand, blind image deblurring methods assume the blur kernel is unknown and aim to simultaneously recover both the sharp image and the blur kernel itself. Since this task is ill-posed, the solution is regularized using various additional constraints~\cite{bahat2017non,cho2009fast,fergus2006removing,xu2010two}.  While these non-deep learning methods show good performance in certain cases, they typically do not perform well in more complicated yet common scenarios such as strong motion blur.

Recent advances of deep learning techniques have revolutionized the field of computer vision; significant progress has been made in numerous domains, including image classification \cite{he2016deep,simonyan2014very}
and  object detection  \cite{he2017mask,isola2017image,ren2015faster,zhu2017unpaired}.  
Image deblurring is no exception: a large number of deep learning methods have been developed for single image and video deblurring, and have advanced the state of the art.  However, the introduction of new  methods with different network designs makes it challenging  to obtain a rapid overview of the field.
This paper aims to fill this gap by providing a survey of recent advances, and to serve as a reference point for new researchers.  

Specifically,  we will focus the discussion on recently published deep learning based image and video deblurring methods.  The aims of this paper are:

\begin{itemize}
\item
To review the preliminaries for image deblurring, including problem definitions, causes of blur, deblurring approaches, quality assessment metrics, and benchmark datasets for performance evaluation.

\item
To discuss new developments of deep learning models for single image and video deblurring and provide a taxonomy for categorizing the existing methods (see Figure~\ref{taxonomy}).

\item
To analyze the challenges of image deblurring and discuss research opportunities.

\end{itemize}

The paper is organized as follows. In Section~\ref{sec_related_work}, we discuss the problem formulation, the causes of blur, the types of deblurring, and image quality metrics. 
Sections~\ref{sec_single_image_non_blind} and \ref{sec_single_image} introduce the CNN-based non-blind and blind image deblurring methods, respectively.
Loss functions applied in deep deblurring methods are discussed in Section~\ref{sec_loss}. 
We introduce public benchmark datasets and evaluations in Sections~\ref{sec_dataset} and~\ref{sec_experiments}, respectively. 
In Section~\ref{sec_domain}, we review three deblurring methods for specific domains, for face, text, and stereo images.
Finally, we discuss the challenges and future opportunities in this research area.

\section{Preliminaries} 
\label{sec_related_work}
\subsection{Problem Formulation}

Image blur can be caused by various factors during image capture: camera shake, in-scene motion, or out-of-focus blur. We denote a blurred image $I_b$ as  
\begin{equation}
\label{f_blur_degradation_1}
I_b = \Phi (I_s; \theta_{\eta}) ,
\end{equation}
where $\Phi$ is the image blur function, and $\theta_{\eta}$ is a parameter vector. $I_s$ is the latent sharp version of the blurred image $I_b$.
Deblurring methods can be categorized into non-blind and blind methods, depending on whether or not the blur function is known (see Sections~\ref{sec_single_image_non_blind} and \ref{sec_single_image}). 
The goal of image deblurring is to recover a sharp image, \ie, finding the inverse of the blur function, as 
\begin{equation}
\label{f_blur_degradation_2}
I_{db} = \Phi^{-1} (I_b; \theta_{\eta}),
\end{equation}
where $\Phi^{-1}$ is the deblurring model, and $I_{db}$ denotes the deblurred image, which is the estimate of the latent sharp image $I_s$.

{\flushleft \textbf{Motion Blur.}}
An image is captured by measuring photons over the time period of camera exposure.
Under bright illumination the exposure time is sufficiently short for the image to capture an instantaneous moment. However, a longer exposure time may result in motion blur.
Numerous methods directly model the degradation process as a convolution process by assuming that the blur is uniform across the entire image: 
\begin{equation}
\label{f_normal}
I_b = K \ast I_s + \theta_{\mu}\,,
\end{equation}
where $K$ is the blur kernel and $\theta_{\mu}$ represents additive Gaussian noise. 
In such an image, any object moving with respect to the camera will look blurred along the direction of relative motion. When we use Eq. \ref{f_normal} to represent the blur process Eq. \ref{f_blur_degradation_1}, $\theta_\eta$   corresponds to a blur kernel and Gaussian noise, while $\Phi$ corresponds to convolution and sum operator. For camera shake, motion blur occurs in the static background, while, in the absence of camera shake, fast moving objects will cause these objects to be  blurred while the background remains sharp. A blurred image can naturally contain blur caused by both factors.
Early methods model blur using shift-invariant kernels \cite{fergus2006removing,xu2014deep}, while more recent studies address the case of non-uniform blur \cite{gao2019dynamic,kupyn2018deblurgan,kupyn2019deblurgan,nah2017deep,tao2018scale}.

{\flushleft \textbf{Out-of-focus Blur.}}
Aside from motion blur, image sharpness is also affected by the distance between the  scene and the camera's focal plane. 
Points on the focal plane are in true focus, and points close to it appear in focus, defining the depth of field.
If the scene contains objects outside this region, parts of the scene will appear blurry. 
The Point Spread Function (PSF) for out-of-focus blur \cite{lu2017out} is often modeled as:
\begin{equation}
\label{eq-defocus}
K(x,y)=\left\{
\begin{aligned}
\frac{1}{\pi r^2}, & \ \ \ \text{if} \ (x-k)^2 + (y-l)^2 \leq r^2, \\
0, & \ \ \ \text{elsewhere},
\end{aligned}
\right.
\end{equation}
where $(k,l)$ is the center of the PSF and $r$ the radius of the blur. 
Out-of-focus deblurring has  applications in saliency detection \cite{jiang2013salient}, defocus magnification \cite{bae2007defocus} and image refocusing \cite{zhang2009single}.
To address the problem of out-of-focus blur, classic methods remove blurry artifacts via blur detection \cite{shi2014discriminative} or coded apertures \cite{masia2011coded}. 
Deep neural networks have been used to detect blur regions \cite{tang2019defusionnet,zhao2019enhancing} and predict depth \cite{anwar2017depth} to guide the deblurring process.

{\flushleft \textbf{Gaussian Blur.}}
Gaussian convolution is a common simple blur model used in image processing, defined as 
\begin{equation}
\label{f_deblur}
G(x,y) = \frac{1}{2\pi \sigma^2} e^{-\frac{x^2+y^2}{2 \sigma^2}},
\end{equation}
where $x$ and $y$ are the distance from the origin in the horizontal and vertical axis, respectively, $\sigma$ is the standard deviation.
Several methods have been developed to remove the Gaussian blur~\cite{chen2009empirical,hummel1987deblurring,vairy1995deblurring}. 

{\flushleft \textbf{Mixed Blur.}}
In many real-world scenes multiple factors contribute to blur, such as camera shake, object motion, and depth variation. 
For example, when a fast-moving object is captured at an out-of-focus distance, the image may include both motion blur and out-of-focus blur as shown in Figure~\ref{idea:d}.
To synthesize this type of blurry image, one option is to firstly transform sharp images to their motion-blurred versions (\textit{e.g.}, by averaging neighboring sharp frames taken in sequence) and then apply an out-of-focus blur kernel based on Eq.~\ref{eq-defocus}. Alternatively, one can train a blurring network to directly generate realistically blurred images \cite{chen2018reblur2deblur,zhang2020deblurring}.

In addition to these main types of blur there can be other causes, such as channel-dependent blur resulting from chromatic aberration \cite{son2011pair,sun2017revisiting}. 

\subsection{Image Quality Assessment}
\label{sec_assessment}
Methods for image quality assessment (IQA) can be classified into subjective and objective metrics. 
Subjective approaches are based on human judgment, which may not require a reference image. One representative metric is the Mean Opinion Score (MOS) \cite{hossfeld2016qoe}, where people rate the quality of images on a scale of 1-5.  As MOS values depend on the population sample, methods typically take the statistics of opinion scores into account.
For image deblurring, most existing methods are evaluated on objective assessment scores, which can be further split into two categories: full-reference and no-reference IQA metrics. 

{\flushleft \textbf{Full-Reference Metrics.}}
Full-reference metrics assess the image quality by comparing the restored image with the ground-truth (GT).
Such metrics include PSNR \cite{hore2010image}, SSIM \cite{wang2004image}, WSNR \cite{mitsa1993evaluation}, MS-SSIM \cite{wang2003multiscale}, IFC \cite{sheikh2005information}, NQM \cite{damera2000image}, UIQI \cite{wang2002universal}, VIF \cite{sheikh2006image}, and LPIPS \cite{Zhang_2018_CVPR}. Among these, PSNR and SSIM are the most commonly used metrics in image restoration tasks \cite{gao2019dynamic,kupyn2018deblurgan,kupyn2019deblurgan,nah2017deep,shen2019human,suin2020spatially,tao2018scale,zhang2018dynamic,zhang2020deblurring}. %
On the other hand, LPIPS and E-LPIPS are able to approximate human judgment of image quality \cite{kettunen2019lpips,Zhang_2018_CVPR}.

{\flushleft \textbf{No-Reference Metrics.}}
While the full-reference metrics require a ground-truth image for evaluation, no-reference metrics use only the deblurred images to measure the quality.
To evaluate the performance of deblurring methods on real-world images, several no-reference metrics have been used, such as BIQI \cite{moorthy2010two}, BLINDS2 \cite{saad2012blind}, BRISQUE \cite{mittal2012no}, CORNIA \cite{ye2012unsupervised}, DIIVINE \cite{moorthy2011blind}, NIQE \cite{mittal2012making}, and SSEQ \cite{liu2014no}.
Further, a number of metrics have been developed to evaluate the performance of image deblurring algorithms by measuring the effect on the accuracy of different vision tasks, such as object detection and recognition \cite{li2018face,yasarla2019deblurring}. 

\begin{table*}[htbp] \scriptsize
	\caption{Overview of deep single image non-blind deblurring methods, where ``Convolution'' denotes convolving sharp images with blur kernels using Eq. \ref{f_normal} to synthesize training data.} 
	\begin{center}\footnotesize{
			\begin{tabular}{M{50pt}|M{30pt}|M{55pt}|M{45pt}|M{45pt}|M{195pt}}
				\toprule
				Method & Category & Blur type & Dataset &  Architecture &  Key idea \\
				\midrule
				
				DCNN \cite{xu2014deep}  & & Gaussian, disk &  & &  first work to combine traditional optimization-based schemes and neural networks \\
				\cline{6-6}
				
				IRCNN \cite{zhang2017learning_deep}  & & Gaussian, motion & & & learn a set of CNN denoisers, use as a modular part of model-based optimization methods to tackle inverse problems \\
				\cline{6-6}
				
				FCNN \cite{zhang2017learning}  & & Motion &  & & adaptively learn image priors to preserve image details and structure with a robust L1 loss \\
				\cline{6-6}
				
				FDN \cite{kruse2017learning} & Uniform & Motion & Convolution & CNN &  learn a CNN-based prior  with an FFT-based deconvolution scheme \\
                \cline{6-6}

				GLRA \cite{ren2018deep} & & Gaussian, disk, motion & & &  use generalized low-rank approximations of blur kernels to initialize the CNN parameters \\ 
				\cline{6-6}
				
				DUBLID \cite{li2019deep} & & Motion &   &  & recast a generalized TV-regularized algorithm into a deep network for blind image deblurring\\
				\cline{6-6}
				
				RGDN \cite{gong2020learning} & & Motion &   & &  incorporate deep neural networks into a fully parameterized gradient descent scheme   \\
				\cline{6-6}
				
				DWDN \cite{dong2021deep} & & Motion, Gaussian &   & & apply explicit deconvolution in  feature space by integrating a classical Wiener deconvolution framework   \\
				\cline{6-6}
				
				USRNet \cite{Zhang_2020_CVPR} & & Motion, Gaussian &   & & end-to-end training of an unfolding network that integrates advantages of model-based and learning-based methods    \\
			
				\bottomrule
		\end{tabular}}
	
		\label{tab-previous-overview_non}
	\end{center}
\end{table*}

\section{Non-Blind Deblurring} 
\label{sec_single_image_non_blind}

The goal of image deblurring is to recover the latent image $I_s$ from a given blurry one $I_b$. If the blur kernel is given, the problem is also known as non-blind deblurring. Even if the blur kernel is available, the task is challenging due to sensor noise and the loss of high-frequency information.

Some non-deep methods employ natural image priors, \eg, global \cite{krishnan2009fast} and local \cite{zoran2011learning} image priors, either in the spatial domain \cite{ren2017video} or in the frequency domain \cite{kruse2017learning} to reconstruct sharp images. To overcome undesired ringing artifacts, Xu~\textit{et al.} \cite{xu2014deep} and Ren~\textit{et al.} \cite{ren2018deep} combine spatial deconvolution and deep neural networks. In addition, several approaches have been proposed to handle saturated regions \cite{cho2011handling,whyte2014deblurring} and to remove unwanted artifacts caused by image noise \cite{kheradmand2014general,nan2020variational}.

We summarize existing deep learning based non-blind methods in Table~\ref{tab-previous-overview_non}.  These approaches can be broadly categorized into two groups: the first group uses deconvolution followed by denoising, while the second group directly employs deep networks.

{\flushleft \textbf{Deconvolution with Denoising.}} Representative algorithms in this category include  \cite{ren2018deep,schuler2013machine,xu2014deep,zhang2017learning,zhang2017learning_deep}. Schuler~\etal~\cite{schuler2013machine} develop a multi-layer perceptron (MLP) to deconvolve images. This approach first recovers a sharp image through a regularized inverse in the Fourier domain, and then uses a neural network to remove artifacts produced in the deconvolution process. 
Xu~\etal~\cite{xu2014deep} use a deep CNN to deblur images containing outliers. This algorithm applies  singular value decomposition to a blur kernel and draws a connection between traditional optimization-based schemes and CNNs. However, the model needs to be retrained for different blur kernels. 
Using the low-rank property of the pseudo-inverse kernel in \cite{xu2014deep}, Ren~\etal~\cite{ren2018deep} propose a generalized deep CNN to handle arbitrary blur kernels in a unified framework without re-training for each kernel.
However, low-rank decompositions of blur kernels can lead to a drop in performance. 
Both methods \cite{ren2018deep,xu2014deep} concatenate a deconvolution CNN and a denoising CNN to remove blur and noise, respectively. However, such denoising networks are designed to remove additive white Gaussian noise and  cannot handle outliers or saturated pixels in blurry images. 
In addition, these non-blind deblurring networks need to be trained for a fixed noise level to achieve good performance, which limits their use in the general case. 
Kruse~\etal~\cite{kruse2017learning} propose a Fourier Deconvolution Network (FDN) by unrolling an iterative scheme, where each stage contains an FFT-based deconvolution module and a CNN-based denoiser. %
Data with multiple noise levels is synthesized for training, achieving better deblurring and denoising performance.

The above methods learn denoising modules for non-blind image deblurring. Learning denoising modules can be seen as learning priors, which will be discussed in the following.

{\flushleft \textbf{Learning priors for deconvolution.}}
Bigdeli~\etal~\cite{bigdeli2017deep} learn a mean-shift vector field representing a smoothed version of the natural image distribution, and use gradient descent to minimize the Bayes risk for non-blind deblurring.
Jin~\etal~\cite{Jin2017Noise} use a Bayesian estimator for simultaneously estimating the noise level and removing blur. They also propose a network (GradNet) to speed up the deblurring process. In contrast to learning a fixed image prior, GradNet can be integrated with different priors and improves existing MAP-based deblurring algorithms.
Zhang~\etal~\cite{zhang2017learning_deep} train a set of discriminative denoisers and integrate them into a model-based optimization framework to solve the non-blind deblurring problem.
Without outlier handling, non-blind deblurring approaches tend to generate ringing artifacts, even in cases when the estimated kernel is accurate. 
Note that some super-resolution methods (with a scale factor of 1) can be adopted for the task of non-blind deblurring as the formulation as image reconstruction task is very similar, \eg, USRNet \cite{Zhang_2020_CVPR}.

\section{Blind Deblurring}
\label{sec_single_image}

In this section, we discuss recent blind deblurring methods. For blind deblurring, both the latent image and the blur kernel are unknown. Early blind deblurring methods focus on removing uniform blur \cite{cho2009fast,fergus2006removing,michaeli2014blind,schuler2015learning,xu2010two}. However, real-world images typically contain non-uniform blur, where different regions in the same image are generated by different blur kernels \cite{rim2020real}. Numerous approaches have been developed to address non-uniform blur by modeling the blur kernel from 3D camera motion \cite{hirsch2011fast,whyte2012non}. Although these approaches can model out-of-plane camera shake, they cannot handle dynamic scenes, which motivated the use of blur fields of moving objects \cite{chakrabarti2010analyzing,gast2016parametric,hyun2013dynamic}. Motion discontinuities and occlusions make the accurate estimation of blur kernels challenging. Recently, several deep learning based methods have been proposed for the deblurring of dynamic scenes \cite{gao2019dynamic,nah2017deep,tao2018scale}.

Tables~\ref{tab-previous-overview} and~\ref{tab-previous-overview_video} summarize representative single image and video deblurring methods, respectively. 
To analyse these methods, we first introduce frame aggregation methods for network input. We then review the basic layers and blocks used in existing  deblurring networks. Finally, we discuss the architectures, as well as the advantages and limitations of current methods.

\subsection{Network inputs and frame aggregation}
Single image deblurring networks take a single blurry image as input and generate the corresponding deblurred result. 
Video deblurring methods take multiple frames as input and aggregate the frame information in either the image or feature domain.
Image-level aggregation algorithms, \eg, Su~\etal~\cite{su2017deep}, stack multiple frames as input and estimate the deblurred result for the central frame. 
On the other hand, feature-level aggregation approaches, \eg, Zhou~\etal~\cite{zhou2019spatio} and Kim~\etal~\cite{hyun2017online}, first extract features from the input frames and then fuse the features for predicting the deblurred results.

\subsection{Basic Layers and Blocks}

This section briefly reviews the most common network layers and blocks used for image deblurring. 

{\flushleft {\textbf{Convolutional layer.}}} Numerous methods \cite{chakrabarti2016neural,kaufman2020deblurring,schuler2015learning,sun2015learning} train 2D CNNs to directly recover sharp images without kernel estimation steps \cite{aljadaany2019douglas,gong2017motion,lu2019unsupervised,madam2018unsupervised,mustaniemi2019gyroscope,nimisha2017blur,shen2019human,zhang2019deep2,zhang2019deep}. 
On the other hand, several approaches use additional prior information, such as depth \cite{li2020dynamic} or semantic labels \cite{shen2018deep}, to guide the deblurring process. 
In addition, 2D convolutions are also adopted by all video deblurring methods \cite{aittala2018burst,hyun2017online,kim2018spatio,nah2019ntire,nah2019recurrent,sim2019deep,su2017deep,wang2019edvr,wieschollek2017learning}. 
The main difference between single image and video deblurring is 3D convolutions, which can extract features from both spatial and temporal domains \cite{zhang2018adversarial}.

\begin{table*}[htbp] \scriptsize
	\caption{Overview of deep single image blind deblurring methods. In the `Dataset' column `averaging' refers to averaging over temporally consecutive sharp frames to synthesize training data.}
	\begin{center}\footnotesize{
			\begin{tabular}{M{50pt}|M{30pt}|M{32pt}|M{45pt}|M{40pt}|M{220pt}}
				\toprule
				Method & Category & Blur type & Dataset & {\scriptsize Architecture} & Key idea  \\
				\midrule

			     Learning-to-Deblur \cite{schuler2015learning} &  & Motion &  & Cascade & The first stage uses a CNN to estimate blur kernels and latent images. The second stage operates on the blurry images and latent image for kernel estimation. \\
		      	\cline{6-6}

			     TextDBN \cite{hradivs2015convolutional} & Uniform & Motion \& defocus & Convolution & CNN & Trains a CNN for blind deblurring and denoising. \\
		      	\cline{6-6}
		      	
				SelfDeblur \cite{ren2019neural} &  & Gaussian \& motion &  & DAE & Two generative networks capture the blur kernel and a latent sharp image, respectively, which is trained on blurry images. \\
			    \midrule
				
				 MRFCNN \cite{sun2015learning} &  &  & Convolution & CNN & Estimate motion kernels from local patches via CNN. An MRF model predicts the motion blur field. \\
				 \cline{6-6}
		      	
				 NDEBLUR \cite{chakrabarti2016neural} &  & & Convolution & CNN & Train a network to generate the complex Fourier coefficients of a deconvolution filter, which is applied to the input patch. \\
				\cline{6-6}
				
			     MSCNN \cite{nah2017deep} & & & Averaging & MS-CNN & A multi-scale CNN generates a low-resolution deblurred image and a deblurred version at the original resolution. \\
		      	\cline{6-6}

				 BIDN \cite{nimisha2017blur} & &  & Convolution & DAE & The network regresses over encoder-features to obtain a blur invariant representation, which is fed into a decoder to generate the sharp image.\\
				\cline{6-6}

				 MBKEN \cite{xu2017motion} & &  & Convolution & Cascade & A two-stage CNN extracts sharp edges from blurry images for kernel estimation. \\
				\cline{6-6}
				
				RNN\_Deblur \cite{zhang2018dynamic} & &  & Convolution & RNN &  Deblurring via a spatially variant RNN, whose weights are learned via a CNN.  \\
				\cline{6-6}

				SRN \cite{tao2018scale} & &  & Averaging & MS-LSTM & Deblurring via a scale-recurrent network that shares network weights across scales. \\
				\cline{6-6}

				DeblurGAN \cite{kupyn2018deblurgan} & Non-uniform & Motion & Averaging & GAN & A conditional GAN-based network generates realistic deblurred images. \\
				\cline{6-6}
				
				UCSDBN \cite{madam2018unsupervised} & & & Convolution & Cycle-GAN &  An unsupervised GAN performs class-specific deblurring using unpaired images as training data. \\
				\cline{6-6}
				
				DMPHN \cite{zhang2019deep2} & &  & Convolution & DAE & A DAE network recovers sharp images based on different patches. \\
				\cline{6-6}
				
				DeepGyro CNN \cite{mustaniemi2019gyroscope} &  &  &  Convolution & DAE & A motion deblurring CNN makes use of the camera's gyroscope readings. \\
				\cline{6-6}
				
				PSS\_SRN \cite{gao2019dynamic}  & &  & Averaging & MS-LSTM & A selective parameter sharing scheme is applied to the SRN architecture and  ResBlocks are replaced by nested skip connections.\\
				\cline{6-6}
				
				DR\_UCSDBN \cite{lu2019unsupervised} &  &  &  Convolution & Cycle-GAN & Unsupervised domain-specific deblurring method by disentangling the content and blur features from input images. \\
				\cline{6-6}
				
                Dr-Net \cite{aljadaany2019douglas} &  &  & Averaging & CNN & A network to learn both the image prior and data fidelity terms via Douglas-Rachford iterations. \\ 
				\cline{6-6}
				
				DeblurGAN-v2 \cite{kupyn2019deblurgan} &  &  & Averaging & GAN & An extension of DeblurGAN using a feature pyramid network and wide range of backbone networks for better speed and accuracy. 
				\\
				\cline{6-6}
	
				RADN \cite{purohit2019region} &  &  &  Averaging & DAE & Region-adaptive dense deformable module to discover spatially varying shifts. \\
				\cline{6-6}
				
				DBRBGAN \cite{zhang2020deblurring} &  &  &  Averaging & Reblur & Two networks, BGAN and DBGAN, which learn to blur and to deblur, respectively.  \\
                \cline{6-6}
                
                SAPHN \cite{suin2020spatially} &  &  &  Averaging & DAE & Content-adaptive architecture to remove spatially-varying image blur.\\
                \cline{6-6}
                
                ASNet \cite{kaufman2020deblurring} &  &  &  Convolution & DAE &  DAE framework, which first estimates the blur kernel in order to recover sharp images. \\
                \cline{6-6}
			
                EBMD \cite{jiang2020learning} &  &  &  Averaging & DAE & An event-based motion deblurring network, introducing a new dataset, DAVIS240C. \\
			
				\bottomrule
		\end{tabular}}
		\label{tab-previous-overview}
	\end{center}
\end{table*}

\begin{table*}[htbp] \scriptsize
	\caption{Overview of deep video deblurring methods. }
	\begin{center}\footnotesize{
			\begin{tabular}{M{50pt}|M{30pt}|M{32pt}|M{45pt}|M{40pt}|M{220pt}}
				\toprule
				Method & Category & Blur type & Dataset & {\scriptsize Architecture} & Key idea\\
				\midrule
				
				DeBlurNet \cite{su2017deep} & &  &  & DAE & Five neighboring blurry images are stacked and fed into a DAE to recover the center sharp image.  \\
				\cline{6-6}

				STRCNN \cite{hyun2017online} & &  &  & RNN-DAE &R ecurrent architecture which includes a dynamic temporal blending mechanism for information propagation. \\
				\cline{6-6}
				
				PICNN \cite{aittala2018burst} & &  &  & DAE &  Permutation invariant CNN which consists of several U-Nets taking a sequence of burst images as input. \\
				\cline{6-6}
				
				DBLRGAN \cite{zhang2018adversarial}  &  &  &  & GAN & GAN-based video deblurring method, using a 3D CNN to extract spatio-temporal information. \\
				\cline{6-6}

				Reblur2deblur \cite{chen2018reblur2deblur} & Non-uniform & Motion & Averaging & Reblur & Three consecutive blurry images are fed into the reblur2deblur framework to recover sharp images, which are used to compute the optical flow and estimate a blur kernel for reconstructing the input. \\
				\cline{6-6}

				IFIRNN \cite{nah2019recurrent} & & & & RNN-DAE &  RNN-based video deblurring network, where a hidden state is transferred from past frames to the current frame. \\
				\cline{6-6}
				
				EDVR \cite{wang2019edvr} & &  &  & DAE & Pyramid, Cascading and Deformable (PCD)  module for frame alignment and a Temporal and Spatial Attention (TSA) fusion module, followed by a reconstruction module to restore sharp videos.  \\
				\cline{6-6}
				
				STFAN \cite{zhou2019spatio} & &  &  & DAE & STFAN takes the current blurred frame, the preceding blurred and restored frames as input and recovers a sharp version of the current frame. \\
				\cline{6-6}

				CDVD-TSP \cite{pan2020cascaded} & &  &  & Cascade &  Cascaded deep video deblurring which first calculates the sharpness prior, then feeds both the blurry images and the prior into an DAE. \\
				\bottomrule
		\end{tabular}}
		\label{tab-previous-overview_video}
	\end{center}
\end{table*}

{\flushleft \textbf{Recurrent layer.}} 
For single image deblurring, recurrent layers can extract features across images at multiple scales in a coarse-to-fine manner. 
Two representative methods are SRN~\cite{tao2018scale} and PSS-SRN \cite{gao2019dynamic}. 
SRN is a coarse-to-fine architecture to remove motion blur via a shared-weight deep autoencoder, while PSS-SRN includes a selective parameter sharing scheme, which leads to improved performance over SRN. 

Recurrent layers can also be used to extract temporal information from neighboring frames in videos \cite{hyun2017online,lumentut2019fast,nah2019recurrent,park2020multi,zhong2020efficient,zhou2019spatio}. The main difference from single image deblurring is that the recurrent layers in video-based methods extract features from neighboring images, rather than transferring information across only one input image at different scales. 
These methods can be categorized into two groups. The first group transfers the feature maps from the last step to the current network to obtain finer deblurred frames \cite{hyun2017online,nah2019recurrent}. The second type of methods generates sharp frames via directly inputting deblurred frames from the last step  \cite{zhou2019spatio}.

{\flushleft \textbf{Residual layer.}}
To avoid vanishing or exploding gradients during training, the global residual layers are used to directly connect low-level and high-level layers in the area of image deblurring \cite{kupyn2018deblurgan,nah2017deep,zhang2018gated}. 
One representative method using this architecture is DeblurGAN \cite{kupyn2018deblurgan} where the output of the deblurring network is added to the input image to estimate the sharp image. In this case, the architecture is equivalent to learning global residuals between blurry and sharp images.
The local ResBlock uses local residual layers, similar to the residual layers in ResNet \cite{he2016deep}, and these are widely used in image deblurring networks \cite{gao2019dynamic,hyun2017online,kupyn2018deblurgan,nah2017deep,nimisha2017blur,tao2018scale}. 
Both kinds of residual layers, local and global, are often combined to achieve better performance.

{\flushleft \textbf{Dense layer.}}
Using dense connections can facilitate addressing the gradient vanishing problem, improving feature propagation, and reducing the number of parameters.
Purohit~\etal~\cite{purohit2019region} propose a region-adaptive dense network composed of region adaptive modules to learn the spatially varying shifts in a blurry image. 
These region adaptive modules are incorporated into a densely connected auto-encoder architecture.
Zhang \etal \cite{zhang2020deblurring} and Gao \etal \cite{gao2019dynamic} also apply dense layers to build their deblurring networks in which DenseBlocks are used to replace the CNN layers or ResBlocks.

{\flushleft \textbf{Attention layer.}}
The attention layer can help deep networks focus on the most important image regions for deblurring. 
Shen~\etal~\cite{shen2019human} propose an attention-based deep deblurring method consisting of three separate branches to remove blur from the foreground, the background, and globally, respectively. 
Since image regions containing people are often of the most interest, the attention module detects the location of people to deblur images using the guidance of a human-aware map.
Other methods employ attention to extract better feature maps, \eg, using self-attention to generate a non-locally enhanced feature map \cite{purohit2019region}.

\begin{figure}[!tb]
  \centering
    \includegraphics[width=0.75\linewidth ]{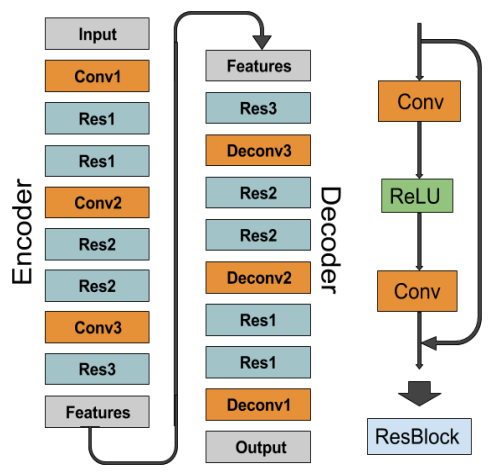}
\caption{Deep single image deblurring network based on the Deep Auto-Encoder (DAE) architecture \cite{nimisha2017blur}. }
  \label{fremework_autoencoder}
\end{figure}

\begin{figure}[!tb]
  \centering
    \includegraphics[width=1\linewidth ]{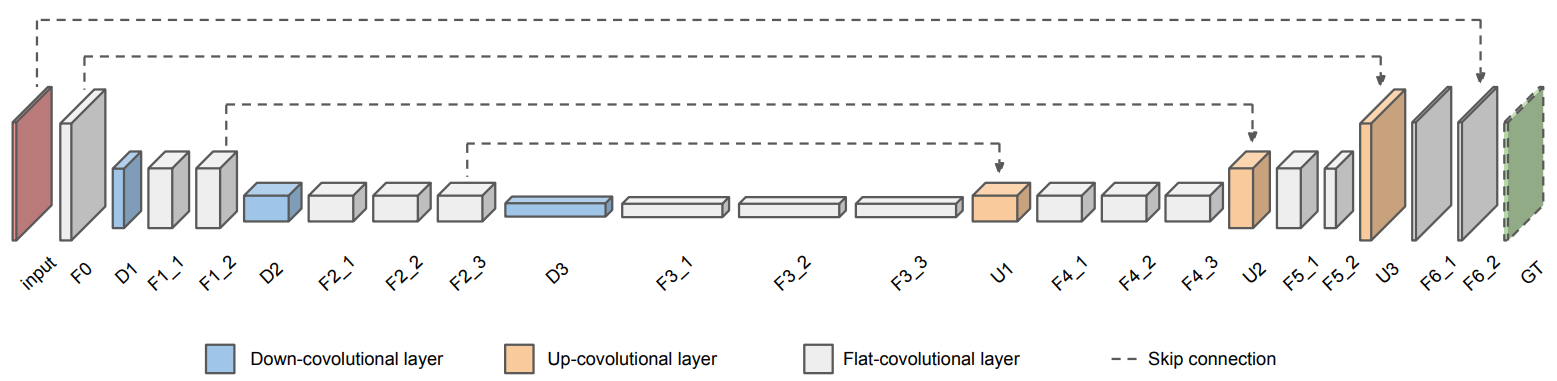}
\caption{Deep video deblurring network based on the DAE architecture \cite{su2017deep}. }
  \label{fremework_autoencoder_video}
\end{figure}

\subsection{Network Architectures} 
\label{network_architectures}
We categorize the most widely used network architectures for image deblurring into five sets: Deep auto-encoders (DAE), generative adversarial networks (GAN), cascaded networks, multi-scale networks, and reblurring networks. We discuss these methods in the following sections.

{\flushleft \textbf{Deep auto-encoders (DAE).}}
A deep auto-encoder first extracts image features and a decoder reconstructs the image from these features.
For single image deblurring, many approaches use the U-Net architecture with a residual learning technique. \cite{gao2019dynamic,nimisha2017blur,shen2018deep,sim2019deep,tao2018scale}. %
In some cases additional networks help exploiting additional information for guiding the U-Net. For example, Shen~\etal~\cite{shen2018deep} propose a face parsing/segmentation network to predict face labels as priors, and use both, blurry images and predicted semantic labels, as the input to a U-Net.
Other methods apply multiple U-Nets to obtain better performance. Tao~\etal~\cite{tao2018scale} analyze different U-Nets as well as DAE, and propose a scale-recurrent network to process blurry images. The first U-net obtains coarse deblurred images, which are  fed into another U-Net to obtain the final result.
The work in \cite{shen2020exploiting} combines the two ideas, using a deblurring network to obtain coarse deblurred images, then feeding them into a face parsing network to generate semantic labels. Finally, both, coarse deblurred images and labels, are fed into a U-Net to obtain the final deblurred images.

\begin{figure}[!tb]
  \centering
    \includegraphics[width=0.99\linewidth ]{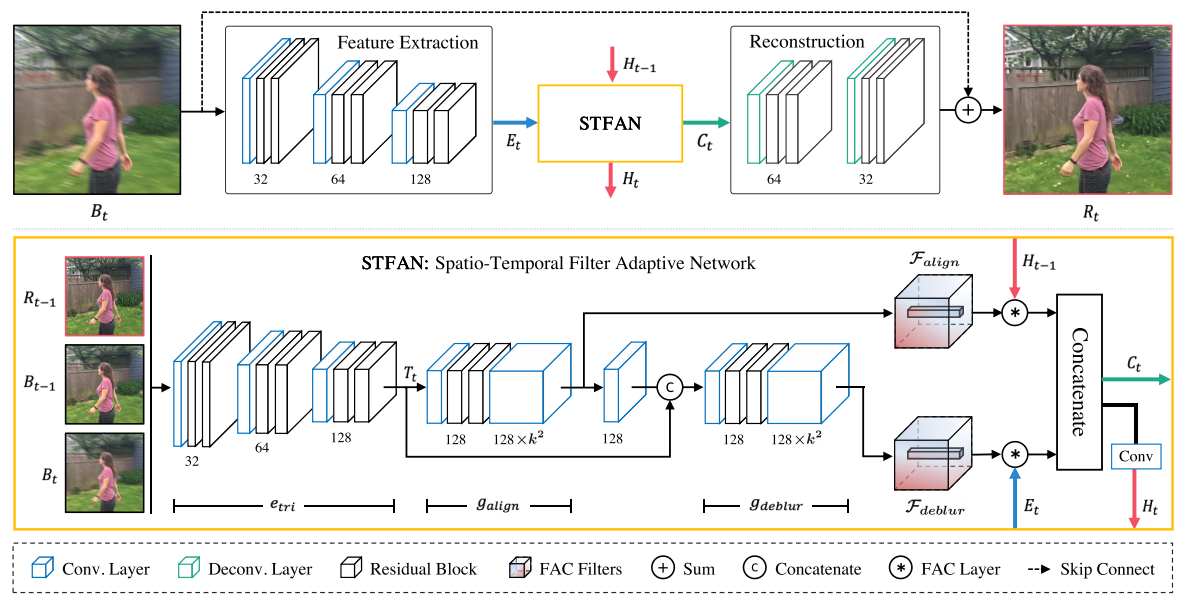}
\caption{Architecture to extract spatio-temporal information based on the STFAN model \cite{zhou2019spatio}. }
  \label{fremework_stfan}
\end{figure}

\begin{figure}[!tb]
  \centering
    \includegraphics[width=0.99\linewidth ]{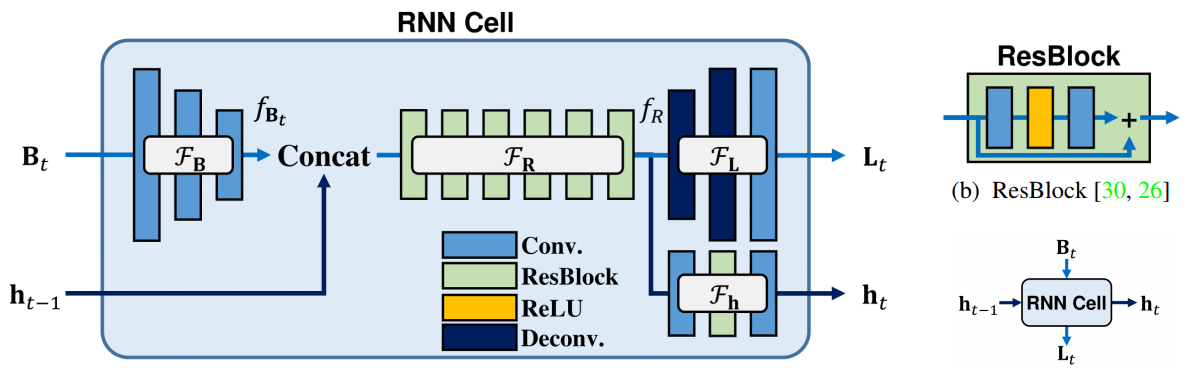}
\caption{Architecture to extract spatio-temporal information based on an RNN model \cite{nah2019recurrent}. }
  \label{fremework_rnn}
\end{figure}

\begin{figure*}[!tb]
  \centering
    \includegraphics[width=0.9\linewidth ]{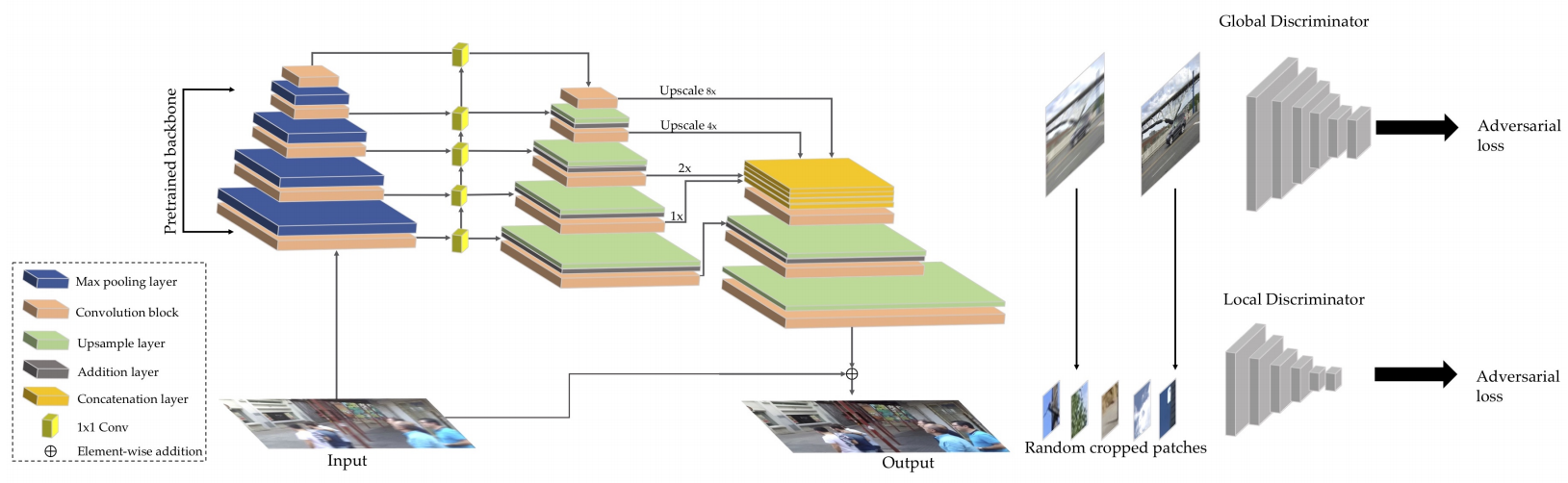}
\caption{Deep single image deblurring network based on the GAN architecture \cite{kupyn2019deblurgan}. }
  \label{fremework_gan}
\end{figure*}

Video deblurring methods can be split into two groups based on their input. The first group takes a stack of neighboring blurry frames as input to extract spatio-temporal information. Su~\etal~\cite{su2017deep} and Wang~\etal~\cite{wang2019edvr} design  DAE architectures to remove blur from videos by feeding several consecutive frames into the encoder, and the decoder recovers the sharp central frame. 
Features extracted from different layers of the encoder are element-wise added to the corresponding decoder layers as shown in Figure~\ref{fremework_autoencoder_video}, which accelerates convergence and generates sharper images.  
The second approach is to feed a single blurry frame into an encoder to extract features. Various  modules have been developed to extract features from neighboring frames, which are jointly fed into the decoder to recover the deblurred frame \cite{zhou2019spatio}. The features from neighboring frames can also be fed into the encoder for feature extraction \cite{hyun2017online}. Numerous deep video deblurring algorithms based on this architecture have been developed~\cite{nah2019recurrent,sim2019deep,wang2019edvr}. 
The main differences are in the module for fusing temporal information from neighboring frames, \eg, STFAN \cite{zhou2019spatio} in Figure~\ref{fremework_stfan} and an RNN module \cite{nah2019recurrent} in Figure~\ref{fremework_rnn}.

\begin{figure}[!tb]
  \centering
    \includegraphics[width=1\linewidth ]{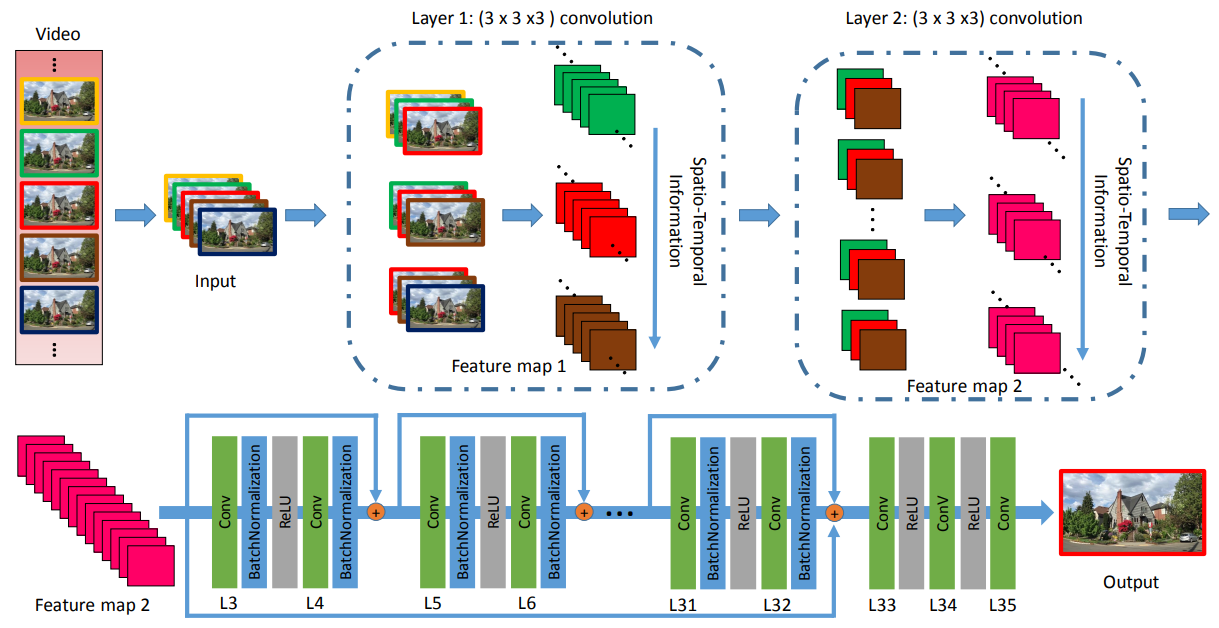}
\caption{Deep video deblurring network based on the GAN architecture \cite{zhang2018adversarial}. }
  \label{fremework_gan_video}
\end{figure}

{\flushleft \textbf{Generative adversarial networks (GAN).}}
GANs have been widely used in image deblurring \cite{kupyn2018deblurgan,kupyn2019deblurgan,nah2017deep,shen2018deep} in recent years. 
Most GAN-based deblurring models share the same strategy: the generator (Figure~\ref{fremework_gan})  generates sharp images such that the discriminator cannot distinguish them from real sharp images.
Kupyn~\etal~\cite{kupyn2018deblurgan} proposed DeblurGAN,  an end-to-end conditional GAN  for motion deblurring. 
The generator of DeblurGAN contains two-strided convolution blocks, nine residual blocks, and two transposed convolution blocks to transform a blurry image to its corresponding sharp version. 
This method is further extended to DeblurGAN-v2 \cite{kupyn2019deblurgan}, which adopts a relativistic conditional GAN and a double-scale discriminator, which consists of
local and global branches as in \cite{isola2017image}.
The core block of the generator is a feature pyramid network, which improves efficiency and performance. 
An adversarial loss is  employed by Nah~\etal~\cite{nah2017deep} and Shen~\etal~\cite{shen2018deep} to generate better deblurred images.

GANs are also used in video deblurring networks. The main difference to single image deblurring is the generator, which also considers  temporal information from neighboring frames.
A 3DCNN model was developed by Zhang~\etal~\cite{zhang2018adversarial} to exploit both spatial and temporal information to restore sharp details using adversarial learning, see Figure~\ref{fremework_gan_video}. 
Kupyn~\etal~\cite{kupyn2019deblurgan} proposed DeblurGAN-v2 to restore sharp videos via modifying the single image deblurring method of DeblurGAN.

\begin{figure}[!tb]
  \centering
    \includegraphics[width=1\linewidth ]{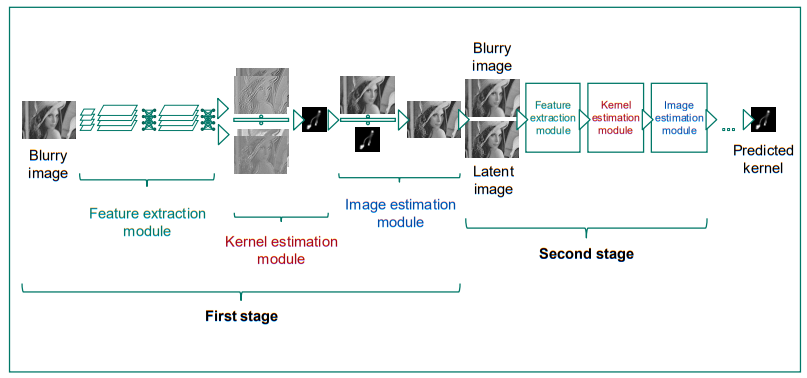}
\caption{Deep network for single image deblurring based on a cascaded architecture \cite{schuler2015learning}. }
  \label{fremework_cascaded}
\end{figure}

{\flushleft \textbf{Cascaded networks.}} A cascaded network contains several modules, which are sequentially concatenated to construct a deeper structure. 
Cascaded networks can be divided into two groups.
The first one uses different architectures in each cascade. 
For example, Schuler~\etal~\cite{schuler2015learning} propose a two-stage cascaded network, see Figure~\ref{fremework_cascaded}. 
The input to the first stage is a blurry image, and the deblurred output is fed into the second stage to predict blur kernels. 
The second group re-trains the same architecture in each cascade to generate deblurred images. Deblurred images from preceding stages are fed into the same type of network to generate finer deblurred results. This cascading scheme can be used in almost all deblurring networks. Although this strategy achieves better performance, the number of CNN parameters increases significantly. To reduce these, recent approaches share parameters in each cascade \cite{pan2020cascaded}.

\begin{figure}[!tb]
  \centering
    \includegraphics[width=1\linewidth ]{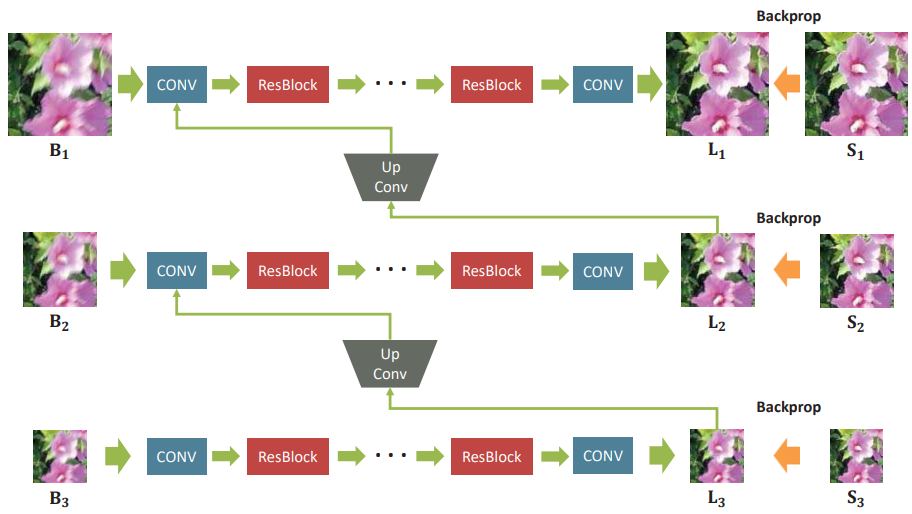}
\caption{Deep network for single image deblurring based on a multi-scale architecture \cite{nah2017deep}.}
  \label{fremework_multiscale}
\end{figure}

\begin{figure}[!tb]
  \centering
    \includegraphics[width=0.99\linewidth ]{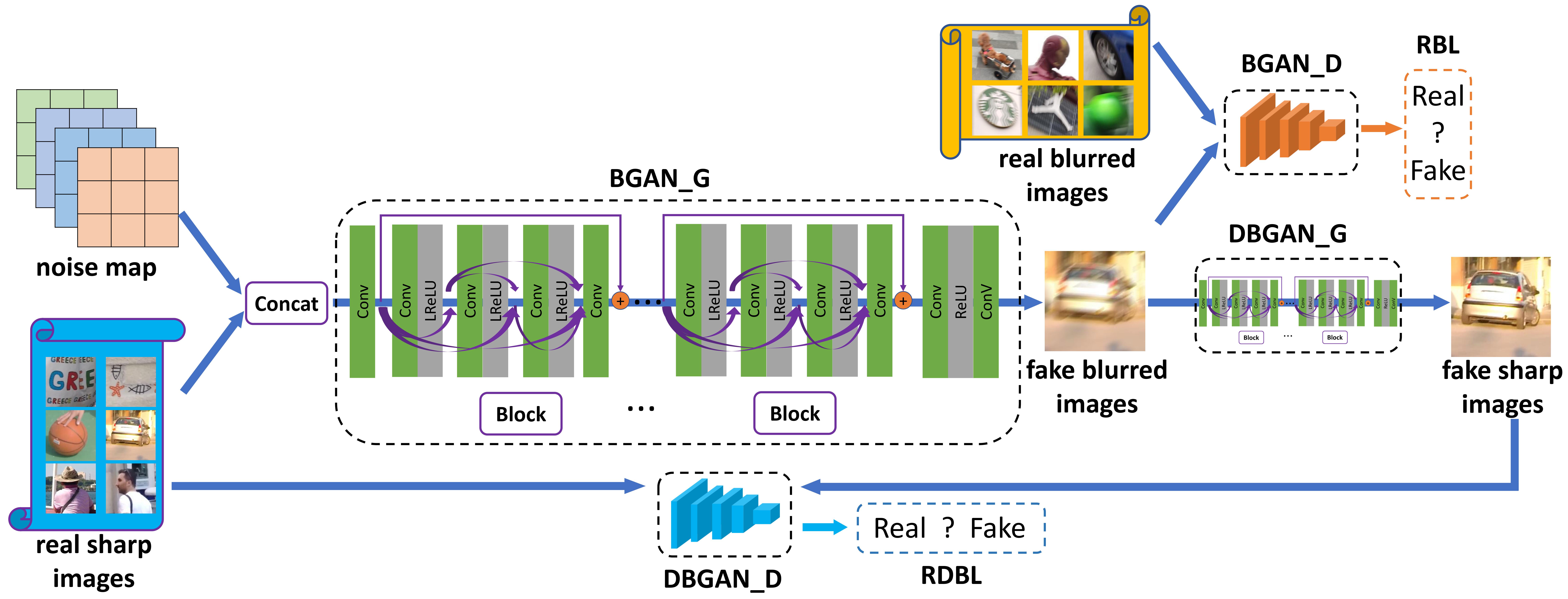}
\caption{Deep network for single image deblurring based on a reblurring architecture \cite{zhang2020deblurring}. }
  \label{fremework_reblurring}
\end{figure}

{\flushleft \textbf{Multi-scale networks.}}
Different scales of the input image describe complementary information \cite{denton2015deep,eigen2014depth,xia2016zoom}. 
The strategy of multi-scale deblurring networks is to first recover low-resolution deblurred images and then progressively generate high-resolution sharp results.
Nah~\etal~\cite{nah2017deep} propose a multi-scale deblurring network to remove motion blur from a single image (Figure~\ref{fremework_multiscale}). 
In a coarse-to-fine scheme, the proposed network first generates images at 1/4 and 1/2 resolutions before estimating the deblurred image at the original scale.
Numerous deep deblurring methods use a  multi-scale architecture \cite{gao2019dynamic,purohit2019bringing,tao2018scale}, improving  deblurring  at different scales, \eg, using nested skip connections in \cite{gao2019dynamic},  or increasing the connection of  networks across different scales, \eg, recurrent layers in \cite{tao2018scale}.

{\flushleft \textbf{Reblurring networks.}}
Reblurring networks can synthesize additional blurry training images \cite{bahat2017non,zhang2020deblurring}. 
Zhang~\etal~\cite{zhang2020deblurring} propose a framework which includes a learning-to-blur GAN (BGAN) and a learning-to-deblur GAN (DBGAN).
This approach learns to transform a sharp image into a blurry version and recover a sharp image from the blurry version, respectively. 
Chen~\etal~\cite{chen2018reblur2deblur} introduce a reblur2deblur framework for video deblurring. Three consecutive blurry images are fed into the framework to recover sharp images, which are further used to compute the optical flow and estimate the blur kernel for reconstructing the blurry input.

{\flushleft {\textbf{Advantages and drawbacks of various models.}}} The U-Net architecture has shown to be effective for image deblurring \cite{shen2018deep,tao2018scale} and low-level vision problems.
Alternative backbone architectures for effective image deblurring include a cascade of Resblocks \cite{kupyn2018deblurgan} or Denseblocks \cite{zhang2020deblurring}.
After selecting the backbone, deep models can be improved in several ways. A multi-scale network \cite{nah2017deep} removes blur at a different scales in a coarse-to-fine fashion, but at increased computational cost~\cite{zhang2019deep}. 
Similarly,  cascaded networks \cite{shen2020exploiting} recover higher-quality deblurred images via multiple deblurring stages. Deblurred images are forwarded to another network to further improve the quality. 
The main difference is that the deblurred images in multi-scale networks are intermediate results, whereas the output of each deblurring network in the cascaded architecture can be individually regarded as final deblurred output. 
Multi-scale architectures and cascaded networks can also be combined by treating the multi-scale networks as a single stage in a cascaded network.

The primary goal of deblurring is to improve the image quality, which may be measured by reconstruction metrics such as PSNR and SSIM. 
However, these metrics are not always consistent with human visual perception. 
GANs can be trained to generate deblurred images, which are deemed realistic according to a discriminator network~\cite{kupyn2018deblurgan,kupyn2019deblurgan}.
For inference, only the generator is necessary. GAN-based models typically perform poorer in terms of distortion metrics such as PSNR or SSIM.

When the number of training samples is insufficient, a reblurring network may be used to generate more data~\cite{chen2018reblur2deblur,zhang2020deblurring}. 
This architecture consists of a \textit{learn-to-blur} and \textit{learn-to-deblur} module.
Any of the above-discussed deep models can be used to synthesize more training samples, creating training pairs of original sharp images and the output from the \textit{learn-to-blur} model.
Although this network can synthesize an unlimited number of training samples, it only models those blur effects that exist in the training samples.

The issue of pixel misalignment is challenging for image deblurring with multi-frame input. 
Correspondences are constructed by computing pixel associations in consecutive frames using  optical flow or geometric transformations. 
For example, a pair of noisy and blurry images can be used for image deblurring and patch correspondences via optical flow \cite{Gu_etal_blur}. 
When applying deep networks, alignment and deblurring can be handled jointly by providing multiple images as input and processing them via 3D convolutions \cite{zhang2018adversarial}.

\section{Loss Functions}
\label{sec_loss}

Various loss functions have been proposed to train deep deblurring networks.  The pixel-wise content loss function has been widely used in the early deep deblurring networks to measure the reconstruction error. However, pixel loss cannot accurately measure the quality of deblurred images. This inspired the development of other loss functions like task-specific loss and adversarial loss for reconstructing more realistic results. In this section, we will review these loss functions.

\subsection{Pixel Loss}
The pixel loss function computes the pixel-wise difference between the deblurred image and the ground truth. The two main variants are the mean absolute error (L1 loss) and the mean square error (L2 loss), defined as:
\begin{equation}
{\mathcal{L}_{pix1}} = \frac{1}{{WH}}\sum\limits_{x = 1}^W {\sum\limits_{y = 1}^H {{| I_{s(x,y)} - I_{db(x,y)}}}| }\, ,
\end{equation}
\begin{equation}
{\mathcal{L}_{pix2}} = \frac{1}{{WH}}\sum\limits_{x = 1}^W {\sum\limits_{y = 1}^H {{{(I_{s(x,y)} - I_{db(x,y)})}^2}} }\, ,
\end{equation}
where $I_{s(x, y)}$ and $I_{db(x, y)}$ are the values of the sharp image and the deblurred image at location $\left(x,y\right)$, respectively.
The pixel loss guides deep deblurring networks to generate sharp images close to the ground-truth pixel values. Most existing deep deblurring networks \cite{chakrabarti2016neural,nah2017deep,tao2018scale,xu2017learning,zhang2018dynamic,zhang2018adversarial} apply the L2 loss since it leads to a high PSNR value. 
Some models are trained to optimize the L1 loss \cite{nimisha2017blur,xu2017motion,zhang2020deblurring}. However, since the pixel loss function ignores long-range image structure, models trained with this loss function tend to generate over-smoothed results \cite{kupyn2019deblurgan}.

\subsection{Perceptual Loss}
Perceptual loss functions \cite{johnson2016perceptual} have also been used to calculate the difference between images. Different from the pixel loss function, the perceptual loss compares the difference in high-level feature spaces such as the features of deep networks trained for classification, \eg, VGG19 \cite{simonyan2014very}. The loss is defined as:
\begin{equation}
{\mathcal{L}_{per}} = \frac{1}{{WHC}} \sqrt{\sum\limits_{x = 1}^W {\sum\limits_{y = 1}^H {\sum\limits_{c = 1}^C {{{(\Phi_{(x,y,c)}^{l} (I_{s}) - \Phi_{(x,y,c)}^{l} (I_{db}))}^2}}}} }\, ,
\end{equation}
where $\Phi_{x,y,c}^{l}(\cdot)$ are the output features of the classifier network from the $l$-th layer. $C$ is the number of channels in the $l$-th layer.
The perceptual loss function compares network features between sharp images and their deblurred versions, rather than directly matching values of each pixel, yielding visually pleasing results \cite{kupyn2018deblurgan,kupyn2019deblurgan,madam2018unsupervised,zhang2018adversarial,zhang2020deblurring}.

\subsection{Adversarial Loss}

For GAN-based deblurring networks, a generator network $G$ and a discriminator network $D$ are trained jointly such that samples generated by $G$ can fool $D$. The process can be modeled as a min-max optimization problem with value function $V(G, D)$:
\begin{equation}
\begin{array}{l}
\mathop {\min }\limits_G \mathop {\max }\limits_D V(G,D) = {{\rm E}_{I \sim {p_{train(I)}}}}[\log (D(I))] \, + \\
\ \ \ \ \ \ \ \ \ \ \ \ \ \ \ \ \ \ \ \ \ {{\rm E}_{I_b \sim {p_{G(I_b)}}}}\left[\log (1 - D(G(I_b)))\right]\, ,
\end{array}
\end{equation}
where $I$ and $I_b$ represent the sharp and blurry images, respectively.
To guide $G$ to generate photo-realistic sharp images, the adversarial loss function is used:
\begin{equation}
{\mathcal{L}_{adversarial}} = \log (1 - D(G({I_{b}})))\, ,
\end{equation}
where $D(G({I_{b}}))$ is the probability that the deblurred image is real.
GAN-based deep deblurring methods have been applied to single image and video deblurring \cite{kupyn2018deblurgan,kupyn2019deblurgan,nah2017deep,shen2018deep,zhang2018adversarial}. In contrast to the pixel and perceptual loss functions, the adversarial loss directly predicts whether the deblurred images are similar to real images and leads to photo-realistic sharp images.

\subsection{Relativistic Loss}

The relativistic loss is related to the adversarial loss, which can be formulated as:
\begin{equation}
\centering
\label{RBL_goal}
\begin{aligned}
D({I_{s}}) & = \sigma (C({I_{s}})) \rightarrow 1 ,
\\ 
D({I_{db}}) & = D(G({I_{db}})) = \sigma (C(G({I_{db}})))  \rightarrow 0 \,,
\end{aligned}
\end{equation}
where $D(\cdot)$ is the probability that the input is a real image, $C(\cdot)$ is the feature captured via a discriminator and $\sigma(\cdot)$ is the activation function. During the training stage, only the second part of Eq. \ref{RBL_goal} updates parameters of the generator $G$, while the first part only  updates the discriminator. 

To train a better generator, the relativistic loss  was proposed to calculate whether a generated image is more realistic than the synthesized images \cite{jolicoeur2018relativistic}. 
Based on this loss function, Zhang~\etal~\cite{zhang2020deblurring} replace the standard adversarial loss with the relativistic loss:
\begin{equation}
\label{RBL_goal_2}
\begin{array}{l}
\sigma (C({I_{s}}) - E(C(G({I_{b}})))) \rightarrow 1\, ,
\\ 
\\
\sigma (C(G({I_{b}})) - E(C({I_{s}}))) \rightarrow 0\,,
\end{array}
\end{equation} 
where $E(\cdot)$ denotes the averaging operation over images in one batch. $C(\cdot)$ is the feature captured via a discriminator and $\sigma(\cdot)$ is the activation function. The generator is updated by the relativistic loss function
\begin{equation}
\small
\begin{array}{l}
{\mathcal{L}}_{RDBL} = -[\log( \sigma(C({I_{r}})-E(C(G(I_{b}))))) 
\\
\\
\ \ \ \ \ \ \ \ \ \ \ \ \ \ \ \ \ \ + \log (1 - (\sigma({G(I_{b})})-E(C(I_{s})))))] , \
\end{array}
\end{equation}
where $I_r$ denotes a real image. The method provided in \cite{kupyn2019deblurgan} also applies this loss function to improve the performance of image deblurring.

\subsection{Optical Flow Loss}

Since the optical flow is able to represent the motion information between two neighboring frames, several studies remove motion blur via estimating optical flow. 
Gong~\etal~\cite{gong2017motion} build a CNN to first estimate the motion flow from the blurred images and then recover the deblurred images based on the estimated  flow field. To obtain pairs of training samples, they simulate motion flows to generate blurred images. 
Chen~\etal~\cite{chen2018reblur2deblur} introduce a reblur2deblur framework, where three consecutive blurry images are input to the deblur sub-net. Then optical flow between the three deblurred images is used to estimate the blur kernel and reconstruct the input.

{\flushleft \textbf{Advantages and drawbacks of different loss functions.}}
Generally, all the above loss functions contribute to the progress of image deblurring. However, their characteristics and goals differ. The pixel loss  generates deblurred images which are close to the sharp ones in terms of pixel-wise measurements. Unfortunately, this typically causes over-smoothing. The perceptual loss is more consistent with human perception, while the results  still exhibit obvious gaps with respect to real sharp images. The adversarial loss and optical flow loss functions aim to generate realistic deblurred images and model the motion blur, respectively. However, they cannot effectively improve the values of PSNR/SSIM or only work on motion blurred images. 
Multiple loss functions can also be used as a weighted sum, trading off their different properties.

\section{Benchmark Datasets for Image Deblurring}
\label{sec_dataset}

In this section, we introduce public datasets for image deblurring.
High-quality datasets should reflect all the different types of blur in real-world scenarios.
We also introduce domain-specific datasets, \eg, face and text images, applicable for domain-specific deblurring methods.
Table \ref{table_dataset} presents an overview of these datasets.

\begin{table*}[tb]
  \centering 
    \caption{Representative benchmark datasets for evaluating single image, video and domain-specific deblurring algorithms. 
    }
    \begin{tabular}{l | c c r r l l r }
    \toprule
    Dataset & Synthetic & Real &  Sharp Images & Blurred Images & Blur Model & Type & Train/Test Split \\
    \midrule
    Levin~\etal~\cite{levin2009understanding} & $\times$ & $\checkmark$ & 4 & 32 & Uniform & Single image & Not divided \\
    Sun~\etal~\cite{sun2012super} & $\checkmark$ & $\times$ & 80 & 640 & Uniform & Single image & Not divided \\
    K\" ohler~\etal~\cite{kohler2012recording} & $\checkmark$ & $\times$ & 4 & 48 & Non-uniform & Single image & Not divided \\
    Lai~\etal~\cite{lai2016comparative} & $\checkmark$ & $\checkmark$ & 108 & 300 & Both & Single image & Not divided \\
    GoPro\cite{nah2017deep} & $\checkmark$ & $\checkmark$ & 3,214 & 3,214 & Non-uniform & Single image & 2,103/1,111\\
    HIDE \cite{shen2019human} & $\checkmark$ & $\times$ & 8,422 & 8,422 & Non-uniform & Single image & 6,397/2,025 \\ 
    Blur-DVS \cite{jiang2020learning} & $\checkmark$ & $\checkmark$ & 2,178 & 2,918 & Non-uniform & Single image & 1,782/396 \\
    \midrule
    Su~\etal~\cite{su2017deep} & $\checkmark$ & $\checkmark$ & 6,708 & 11,925 & Non-uniform & Video & 5,708/1,000 \\   
    REDS \cite{nah2019ntire} & $\checkmark$ & $\checkmark$ & 30,000 & 30,000 & Non-uniform & Video & 24,000/3,000\\   
    \midrule
    Hradi{\v{s}}~\etal~\cite{hradivs2015convolutional} & $\checkmark$ & $\times$ & 3M+35K & 3M+35K & Non-uniform & Text & 3M/35K \\   
    Shen~\etal~\cite{shen2018deep} & $\checkmark$ & $\times$ & 6,564 & 130M+16K & Uniform & Face & 130M/16K \\ 
    Zhou~\etal~\cite{zhou2019davanet} & $\checkmark$ & $\times$ & 20,637 & 20,637 & Non-uniform & Stereo & 17,319/3,318 \\ 
    \bottomrule
    \end{tabular}%
    \label{table_dataset}
\end{table*}%

\subsection{Image Deblurring Datasets}
\label{dataset_single_image}

{\flushleft \textbf{Levin \etal~dataset.}} 
To construct an image deblurring dataset, Levin~\etal~\cite{levin2009understanding} mount the camera on a tripod to capture blur of actual camera shake by locking the Z-axis rotation handle and allowing motion in X and Y-directions. A dataset containing 4 sharp images of size $255 \times 255$ and 8 uniform blur kernels is captured using this set-up.

{\flushleft \textbf{Sun~\etal~dataset.}}
Sun~\etal~\cite{sun2013edge} extend the dataset of Levin~\etal~\cite{levin2009understanding} by using $80$ high-resolution natural images from \cite{sun2012super}.
Applying the 8 blur kernels from \cite{levin2009understanding}, this results in $640$ blurred images. 
Similar to Levin~\etal~\cite{levin2009understanding}, this dataset contains only uniformly blurred images and is insufficient for training robust CNN models.

{\flushleft \textbf{K\" ohler~\etal~dataset.}}
To simulate non-uniform blur, K\" ohler~\etal~\cite{kohler2012recording} use a Stewart platform (\ie, a robotic arm) to record the 6D camera motion and capture a printed picture.
There are $4$ latent sharp images and 12 camera trajectories, resulting in a total of $48$ non-uniformly blurred images. 

{\flushleft \textbf{Lai~\etal~dataset.}} Lai~\etal~\cite{lai2016comparative} provide a dataset which includes 100 real and 200 synthetic blurry images generated using both uniform blur kernels and 6D camera trajectories.
The images in this dataset cover various scenarios, \eg, outdoor, face, text, and low-light images, and thus can be used to evaluate deblurring methods in a variety of settings.

{\flushleft \textbf{GoPro dataset.}}
Nah~\etal~\cite{nah2017deep} created a large-scale dataset to simulate  real-world blur by frame averaging. A motion blurred image can be generated by integrating multiple instant and sharp images over a time interval:
\begin{equation}
\label{f_real_blur}
I_b  = g \left(\frac{1}{T}{\int_{t=0}^{T}I_{s(t)}\mathrm{d}t} \right), 
\end{equation}
where $g(\cdot)$ is the Camera Response Function (CRF), and $T$ denotes the period of camera exposure. 
Instead of modeling the convolution kernel,  $M$ consecutive sharp frames are averaged to generate a blurry image:
\begin{equation}
\label{blurr_process2}
I_b \simeq g \left( \frac{1}{M}\sum_{t=0}^{M-1}I_{S[t]} \right).
\end{equation}
Sharp images, captured at 240fps using a GoPro Hero4 Black camera, are averaged over time windows of varying duration to synthesize blurry images.
The sharp image at the center of a time window is used as ground truth image. The GoPro dataset has been widely used to train and evaluate deep deblurring algorithms. The dataset contains $3,214$ image pairs, split into training and test sets, containing $2,103$ and $1,111$ pairs, respectively.

{\flushleft  \textbf{HIDE dataset.}} Focusing mainly on pedestrians and street scenes, Shen~\etal~\cite{shen2019human} created a  motion blurred dataset, which includes camera shake and object movement. The dataset includes $6,397$ and $2,025$ pairs for training and testing, respectively. Similar to the GoPro dataset \cite{nah2017deep}, the blurry images in the HIDE dataset are synthesized by averaging $11$ continuing frames, where the central frame is used as the sharp image.

{\flushleft  \textbf{RealBlur dataset.}} To train and benchmark deep deblurring methods on real blurry images, Rim \textit{et al.} \cite{rim2020real} created the RealBlur dataset, consisting of two subsets. The first is RealBlur-J, which contains camera JPEG outputs.
The second is RealBlur-R, which contains RAW images. 
The RAW images are generated by using white balance, demosaicking, and denoising operations. The dataset totally contains 9476 pairs of images.

{\flushleft \textbf{Blur-DVS dataset}.}
To evaluate the performance of event-based deblurring methods \cite{jiang2020learning,lin2020learining}, Jiang~\etal~\cite{jiang2020learning} create a Blur-DVS dataset using a DAVIS240C camera. 
An image sequence is first captured with slow camera motion and used to synthesize $2,178$ pairs of blurred and sharp images by averaging seven neighboring frames. Training and test sets contain $1,782$ and $396$ pairs, respectively. The dataset also provides 740 real blurry images without sharp ground-truth images.

\subsection{Video Deblurring Datasets}
\label{dataset_video}

\textbf{DVD dataset.} Su~\etal~\cite{su2017deep} captured blurry video sequences with cameras on different devices, including an iPhone 5s, a Nexus 4x, and a GoPro Hero 4.
To simulate realistic motion blur,
they captured sharp videos at 240 fps and averaged eight neighboring frames to create the corresponding blurry videos at 30fps.
The dataset consists of a quantitative subset and a qualitative subset. The quantitative subset includes $6,708$ blurry images and their corresponding sharp images from $71$ videos (61 training videos and 10 test videos). 
The qualitative subset covers $22$ videos with a $3-5$ second duration, which are used for visual inspection.

{\flushleft \textbf{REDS dataset.}} In order to capture the motion of fast moving objects, 
Nah \textit{et al.} \cite{nah2019ntire} captured $300$ video clips at $120$ fps and  $1080 \times 1920$ resolution using a GoPro Hero 6 Black camera, and  increased the frame rate to $1920$ fps by recursively applying frame interpolation \cite{niklaus2017video}.
By generating $24$ fps blurry videos from the $1920$ fps videos, spike and step artifacts present in the DVD dataset \cite{su2017deep} are reduced.
Both the synthesized blurry frames and the sharp frames are downscaled to $720 \times 1280$ to suppress noise and compression artifacts. This dataset has also been used for evaluating single image deblurring methods \cite{nah2020ntire}.

\subsection{Domain-Specific Datasets}
\label{dataset_domain}

{\flushleft  \textbf{Text deblurring datasets.}} Hradi{\v{s}}~\etal~\cite{hradivs2015convolutional}~collected text documents from the Internet. These are downsampled to $120-150$ DPI resolution, and split into training and test sets of sizes 50K and 2K, respectively. 
Small geometric transformations with bicubic interpolation are also applied to images. Finally, 3M and 35K patches are cropped from the 50K and 2K images respectively for training and testing deblurred models. 
Motion blur and out-of-focus blur are used to generate the blurred images from the sharp ones. 
Cho~\etal~ \cite{cho2012text} also provide a text deblurring dataset with only limited number of images available.

{\flushleft  \textbf{Face deblurring datasets.}} Blurred face image datasets have been constructed from existing face image datasets \cite{shen2018deep,xu2017learning}. 
Shen~\etal~\cite{shen2018deep} collected images from the Helen \cite{le2012interactive} dataset, the CMU PIE \cite{sim2002cmu} dataset, and the CelebA \cite{liu2015deep} dataset. 
They synthesized $20,000$ motion blur kernels to generate 130 million blurry images for training, and used another 80 blur kernels to generate 16,000 images for testing.

{\flushleft \textbf{Stereo blur dataset.}} Stereo cameras are widely used in fast-moving devices such as aerial vehicles and autonomous vehicles. In order to study stereo image deblurring,
Zhou~\etal~\cite{zhou2019davanet} used a ZED stereo camera to capture 60 fps videos, increased to 480 fps via frame interpolation \cite{niklaus2017video}. A varying number of successive sharp images are averaged to synthesize blurry images. This dataset includes 20,637 pairs of images, which are divided into sets of 17,319 and 3,318 for training and testing, respectively.

\begin{figure*}[tb]
  \centering
\includegraphics[width=1\linewidth]{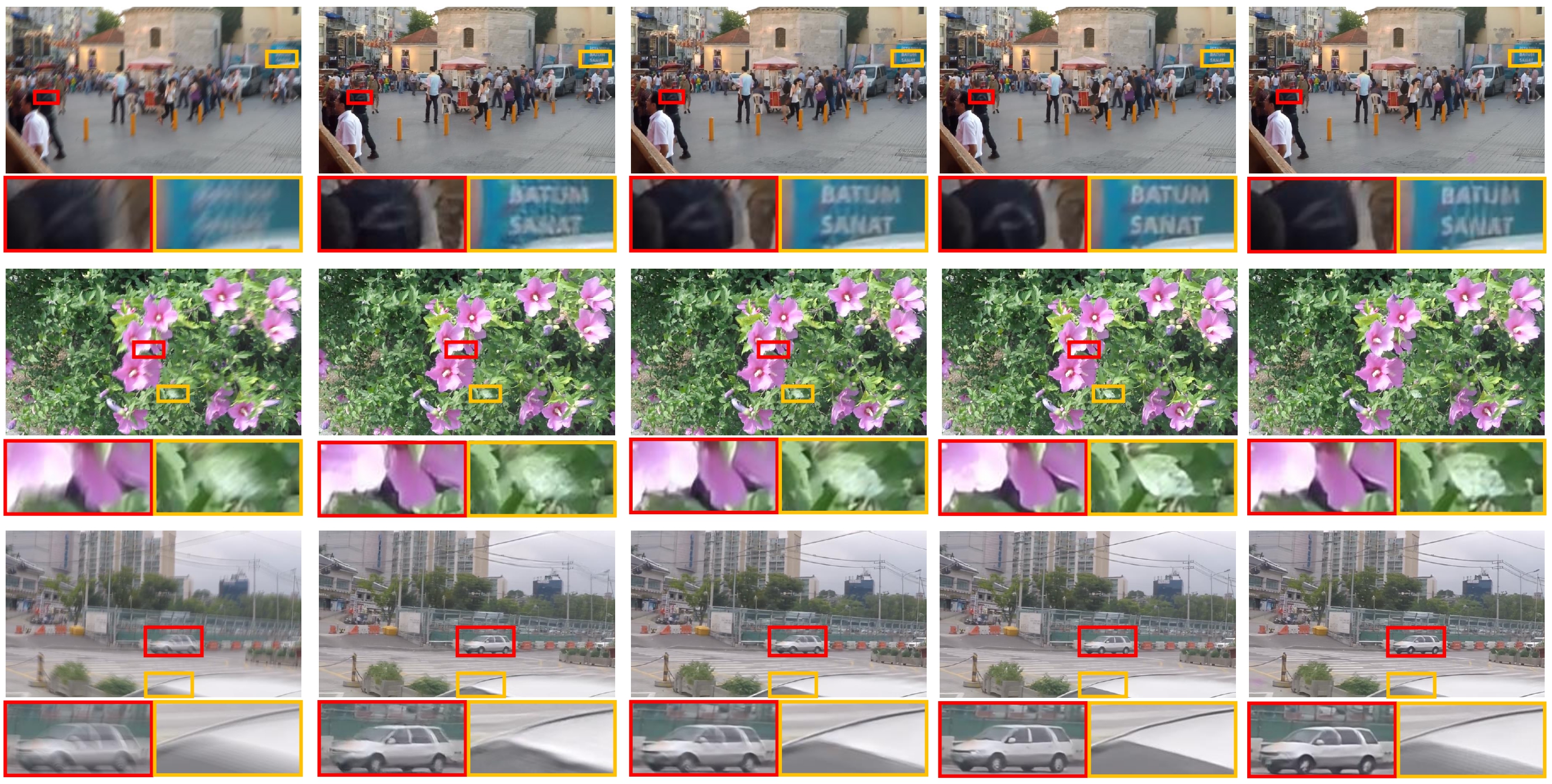}
  \caption{ Evaluation results of the state-of-the-art deblurring methods on the GoPro dataset \cite{nah2017deep}. From left to right: blurry images, results of Nah \etal \cite{nah2017deep}, Tao \etal \cite{tao2018scale}, DBGAN \cite{zhang2020deblurring} and DeblurGAN-v2 \cite{kupyn2019deblurgan}. \cite{nah2017deep} and \cite{tao2018scale} are two multi-scale based image deblurring networks. \cite{zhang2020deblurring} and \cite{kupyn2019deblurgan} are two GAN based image deblurring networks.  }
  \label{ablation_structure}
\end{figure*}

\begin{table*}[tb]
\tiny
  \centering 
    \caption{Performance evaluation of representative methods for single image deblurring on three popular image deblurring datasets. MS stands for  multi-scale.}
    \begin{tabular}{l | l l l l r l  }
    \toprule
    Dataset &  Method  & Framework & Layers \& block &  Loss & PSNR/SSIM & Characteristic \\
    \midrule
    \multirow{11}{*}{GoPro \cite{nah2017deep}} & Sun~\etal~\cite{sun2015learning} & DAE & Convolution & $\mathcal{L}_{pix}$ & 24.64/0.8429 & CNN-based approach, MRF\\
     & Gong~\etal~\cite{gong2017motion} & DAE & Fully convolution & $\mathcal{L}_{pix}$, $\mathcal{L}_{flow}$ & 26.06/0.8632 & Estimation of motion flow\\
     & Nah~\etal~\cite{nah2017deep} & MS, GAN & ResBlock & $\mathcal{L}_{per}$, $\mathcal{L}_{adv}$ & 29.23/0.9162 & Multi-scale Net, adversarial loss\\
     & Kupyn~\etal~\cite{kupyn2018deblurgan} & Conditional GAN & ResBlock & $\mathcal{L}_{per}$, $\mathcal{L}_{adv}$ & 28.70/0.8580 & Conditional GAN-based model\\
    & Zhang~\etal~\cite{zhang2018dynamic} & U-Net & Recurrent, Residual & $\mathcal{L}_{pix}$ & 29.19/0.9323 & Spatially variant RNN \\
     & Tao~\etal~\cite{tao2018scale} & U-Net, MS & Recurrent, Dense & $\mathcal{L}_{pix}$ & 30.26/0.9342 & Scale-recurrent Net \\
     & Gao~\etal~\cite{gao2019dynamic} & U-Net, MS & Recurrent, Dense & $\mathcal{L}_{pix}$ & 31.58/0.9478 & Parameter selective sharing \\
     & Kupyn~\etal~\cite{kupyn2019deblurgan} & Conditional GAN, MS & Residual & $\mathcal{L}_{pix}$, $L_{per}$, $\mathcal{L}_{adv}$ & 29.55/0.9340 & Feature Pyramid Net, fast \\
     & Shen~\etal~\cite{shen2019human} & DAE & Residual, Attention & $\mathcal{L}_{pix}$ & 30.26/0.9400 & Human-aware Net \\
     & Purohit~\etal~\cite{purohit2019region} & U-Net, DAE, MS & Attention, Dense & $\mathcal{L}_{pix}$ & 32.15/0.9560 & Dense deformable module \\
     & Zhang~\etal~\cite{zhang2020deblurring} & Reblurring, GAN & Dense & $\mathcal{L}_{pix}$, $\mathcal{L}_{per}$, $\mathcal{L}_{adv}$ & 31.10/0.9424 & Deblurring via reblurring \\
     & Zhang~\etal~\cite{zhang2019deep2} & Multi-Patch, Cascading & Residual & $\mathcal{L}_{pix}$ & 31.50/0.9483 & Multi-patch Net \\
     & Jiang~\etal~\cite{jiang2020learning} & DAE, Cascading & Convolution, Recurrent & $\mathcal{L}_{pix}$, $\mathcal{L}_{flow}$, $\mathcal{L}_{adv}$ & 31.79/0.9490 & Event-based motion deblurring \\
     & Suin~\etal~\cite{suin2020spatially} & DAE & Convolution, Attention & $\mathcal{L}_{pix}$ & 32.02/0.9530 & Spatially-Attentive Net \\
    \toprule
    \multirow{5}{*}{K\" ohler \cite{kohler2012recording}} & Sun~\etal~\cite{sun2015learning} & DAE & Convolution & $\mathcal{L}_{pix}$ & 25.22/0.773 & CNN-based approach, MRF\\
    & Kupyn~\etal~\cite{kupyn2018deblurgan} & Conditional GAN & ResBlock & $\mathcal{L}_{per}$, $\mathcal{L}_{adv}$ & 26.48/0.807 & Conditional GAN-based model\\
    & Nah~\etal~\cite{nah2017deep} & MS, GAN & ResBlock & $\mathcal{L}_{per}$, $\mathcal{L}_{adv}$ & 26.75/0.837 & Multi-scale Net, adversarial loss\\
    & Tao~\etal~\cite{tao2018scale} & U-Net, DAE, MS & Recurrent, Dense & $\mathcal{L}_{pix}$ & 26.80/0.838 & Scale-recurrent Net \\
    & Kupyn~\etal~\cite{kupyn2019deblurgan} & Conditional GAN, MS & Residual & $\mathcal{L}_{pix}$, $\mathcal{L}_{per}$, $\mathcal{L}_{adv}$ & 26.72/0.836 & Feature Pyramid Net, fast \\
    \toprule
    \multirow{4}{*}{Shen \cite{shen2019human}} & Sun~\etal~\cite{sun2015learning} & DAE & Convolution & $\mathcal{L}_{pix}$ & 23.21/0.797 & CNN-based approach, MRF\\
    & Nah~\etal~\cite{nah2017deep} & MS, GAN & ResBlock & $\mathcal{L}_{per}$, $\mathcal{L}_{adv}$ & 27.43/0.902 & Multi-scale Net, adversarial loss\\
    & Tao~\etal~\cite{tao2018scale} & U-Net, DAE, MS & Recurrent, Dense & $\mathcal{L}_{pix}$ & 28.60/0.941 & Scale-recurrent Net \\
    & Shen~\etal~\cite{shen2019human} & DAE & Residual, Attention & $\mathcal{L}_{pix}$ & 29.60/0.941 & Human-aware Net \\
    \bottomrule
    \end{tabular}%
    \label{table_compare_image}
\end{table*}%

\begin{figure*}[tb]
  \centering
  \tiny
\begin{tikzpicture}
\begin{axis}[
width=18cm,height=7cm, 
symbolic x coords={Sun\cite{sun2015learning}, Gong\cite{gong2017motion}, Nah\cite{nah2017deep}, Kupyn\cite{kupyn2018deblurgan},Zhang\cite{zhang2018dynamic},Tao\cite{tao2018scale}, Gao\cite{gao2019dynamic}, Kupyn\cite{kupyn2019deblurgan}, Shen\cite{shen2019human}, Purohit\cite{purohit2019region}, Zhang\cite{zhang2020deblurring},Jiang\cite{jiang2020learning}, Suin\cite{suin2020spatially}},
xtick=data
]
\addplot[ybar,fill=blue] coordinates { 
    (Sun\cite{sun2015learning},   24.64)
    (Gong\cite{gong2017motion},  26.06)
    (Nah\cite{nah2017deep},   29.23)
    (Kupyn\cite{kupyn2018deblurgan}, 28.70)
    (Zhang\cite{zhang2018dynamic},29.19)
    (Tao\cite{tao2018scale}, 30.26)
    (Gao\cite{gao2019dynamic}, 31.58)
    (Kupyn\cite{kupyn2019deblurgan}, 29.55)
    (Shen\cite{shen2019human}, 30.26)
    (Purohit\cite{purohit2019region}, 32.15)
    (Zhang\cite{zhang2020deblurring}, 31.10)
    (Jiang\cite{jiang2020learning}, 31.79)
    (Suin\cite{suin2020spatially}, 32.02)
};
\end{axis}
\end{tikzpicture}
\caption{Comparison among state-of-the-art image deblurring methods in terms of PSNR on the GoPro dataset.} 
\label{bar_graph_gopro}
\end{figure*}
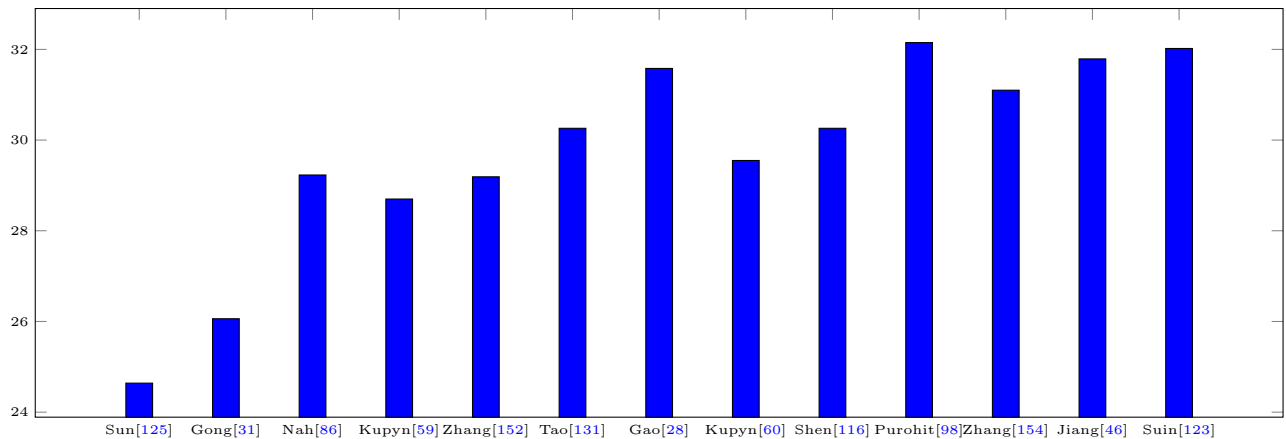

\begin{table}[tb]
  \centering 
    \caption{Performance of representative single deblurring methods using the LPIPS metric on the GoPro dataset. }
    \begin{tabular}{l | c}
    \toprule
    Method & LPIPS \\
    \hline
    Nah~\etal~\cite{nah2017deep}  & 0.1819 \\
    Kupyn~\etal~\cite{kupyn2019deblurgan} & 0.2528 \\
    Tao~\etal~\cite{tao2018scale}  & 0.7879\\
    Gao~\etal~\cite{gao2019dynamic} & 0.0359 \\
    Zhang~\etal~\cite{zhang2020deblurring} & 0.1097 \\ 
     \bottomrule
    \end{tabular}%
    \label{table_compare_image_lpips}
\end{table}%

\begin{table}[tb]
  \centering 
    \caption{Performance evaluation of different loss functions for single image deblurring.}
    \begin{tabular}{l | c c}
    \toprule
    Method & PSNR & SSIM \\
    \hline
    L1  & 26.24 & 0.9012 \\
    L2  & 26.78 & 0.9024 \\
    L1 + Perceptual  & 26.92 & 0.9078 \\
    L2 + Perceptual  & 26.95 & 0.9081 \\
    L1 + Perceptual + GAN  & 26.89 & 0.9060 \\
    L2 + Perceptual + GAN  & 27.20 & 0.9114 \\
    L1 + Perceptual + RaGAN  & 27.19 & 0.9111 \\
    L2 + Perceptual + RaGAN  & 27.09 & 0.9088 \\
     \bottomrule
    \end{tabular}%
    \label{table_ablation_loss_fuction}
\end{table}%

\begin{table}[tb]
  \centering 
    \caption{The NIQE and BRISQUE of representative non-blind and blind methods for single image deblurring on the RWBI dataset. We use public available pre-trained models.}
    \begin{tabular}{l | c c}
    \toprule
    Method & NIQUE & BRISQUE \\
    \hline
    FDN \cite{kruse2017learning}  & 12.4276 & 57.7418 \\
    RGDN \cite{gong2020learning}  & 13.0038 & 51.5109 \\
    \hline
    Nah~\etal~\cite{nah2017deep}  & 12.2365 & 49.9521  \\
    Kupyn~\etal~\cite{kupyn2019deblurgan} & 11.8186 & 40.4656\\
    Tao~\etal~\cite{tao2018scale}  & 12.4606 & 51.1515 \\
    Gao~\etal~\cite{gao2019dynamic} & 12.3987 & 50.5300 \\
    Zhang~\etal~\cite{zhang2020deblurring} & 11.5048  & 45.5496 \\ 
     \bottomrule
    \end{tabular}%
    \label{table_compare_image_niqe_brisque}
\end{table}%

\begin{table}[tb]
  \centering 
    \caption{The PSNR and SSIM values of representative non-blind and blind deblur methods for single image deblurring on the non-blind GoPro dataset. All methods are trained on the non-blind dataset. }
    \begin{tabular}{l | c c c}
    \toprule
    Method & type & PSNR & SSIM \\
    \hline
    DWDN \cite{dong2021deep}  & non-blind & 36.42 & 0.9762  \\
    USRNet \cite{Zhang_2020_CVPR} & non-blind & 36.51 & 0.9823 \\
    Nah~\etal~\cite{nah2017deep} & blind & 33.88 & 0.9795 \\
    SRN \cite{tao2018scale} & blind & 34.78  & 0.9812 \\
     \bottomrule
    \end{tabular}%
    \label{table_compare_image_non_blind}
\end{table}%

\begin{table}[tb]
  \centering 
    \caption{Speed of representative methods for single image deblurring. The numbers are quoted from \cite{suin2020spatially}.}
    \begin{tabular}{l | c}
    \toprule
    Method & Speed (s) \\
    \hline
    Nah~\etal~\cite{nah2017deep}  & 6\\
    Kupyn~\etal~\cite{kupyn2018deblurgan}  & 1\\
    Tao~\etal~\cite{tao2018scale}  & 1.2\\
    Zhang~\etal~\cite{zhang2018dynamic} & 1 \\
    Gao~\etal~\cite{gao2019dynamic}  & 1 \\
    Kupyn~\etal~\cite{kupyn2019deblurgan} & 0.48 \\
    Suin~\etal~\cite{suin2020spatially}  & 0.34 \\
     \bottomrule
    \end{tabular}%
    \label{table_compare_image_speed}
\end{table}%

\begin{figure*}[tb]
  \centering
\includegraphics[width=1\linewidth]{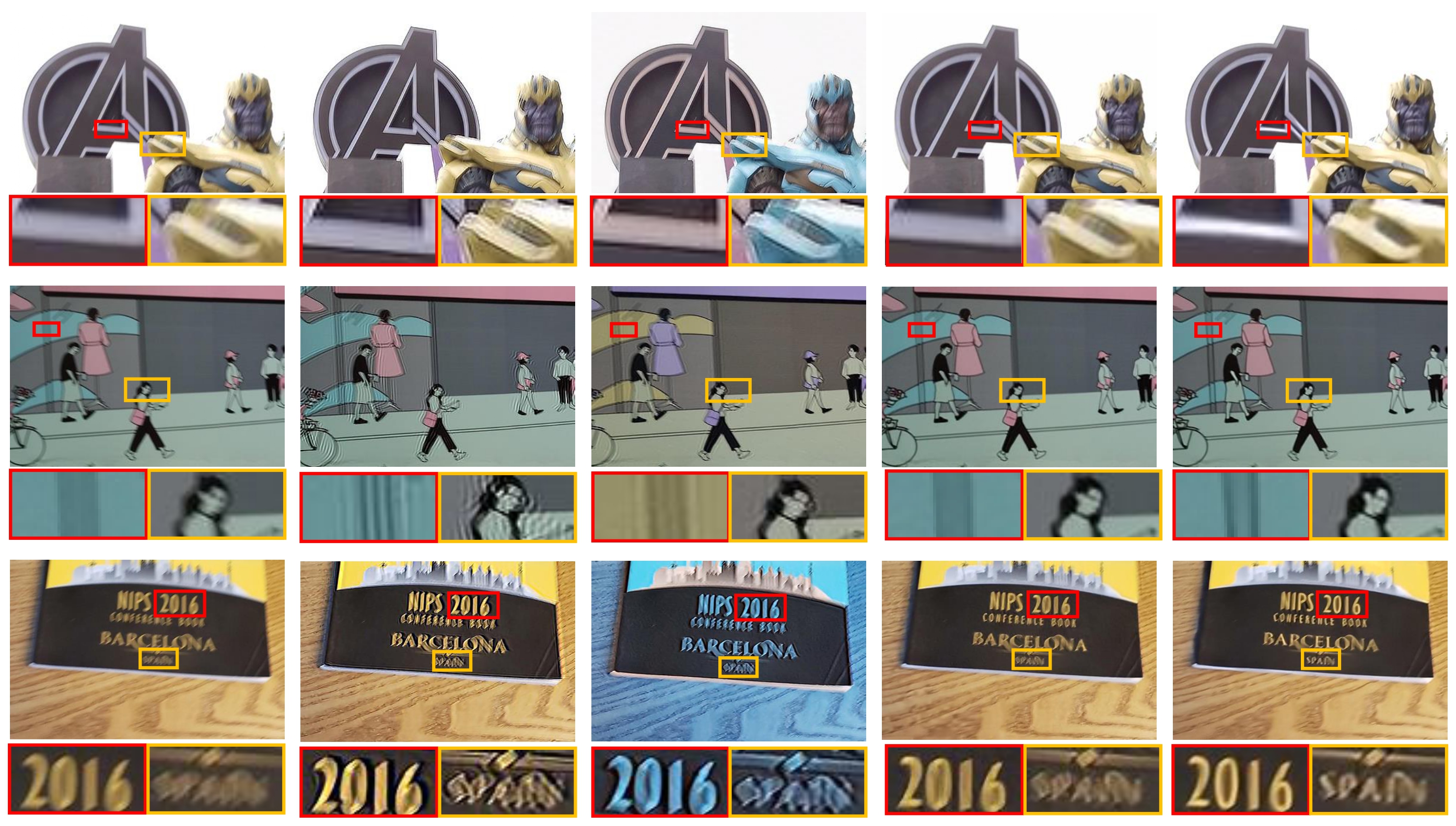}
  \caption{Comparison among state-of-the-art non-blind and blind image deblurring methods. \textcolor{blue}{The blur kernels are predicted by the method in \cite{xu2013unnatural}}. From left to right: blurry images, results of FDN \cite{kruse2017learning}, RGDN \cite{gong2020learning}, SRN \cite{tao2018scale} and DBGAN \cite{zhang2020deblurring}. FDN \cite{kruse2017learning} and RGDN \cite{gong2020learning} are non-blind deblurring methods, whereas SRN \cite{tao2018scale} and DBGAN \cite{zhang2020deblurring} are blind deblurring methods.  } 
  \label{fig_rwbi_blind_non_blind}
\end{figure*}

\begin{figure*}[tb]
  \centering
\includegraphics[width=1\linewidth]{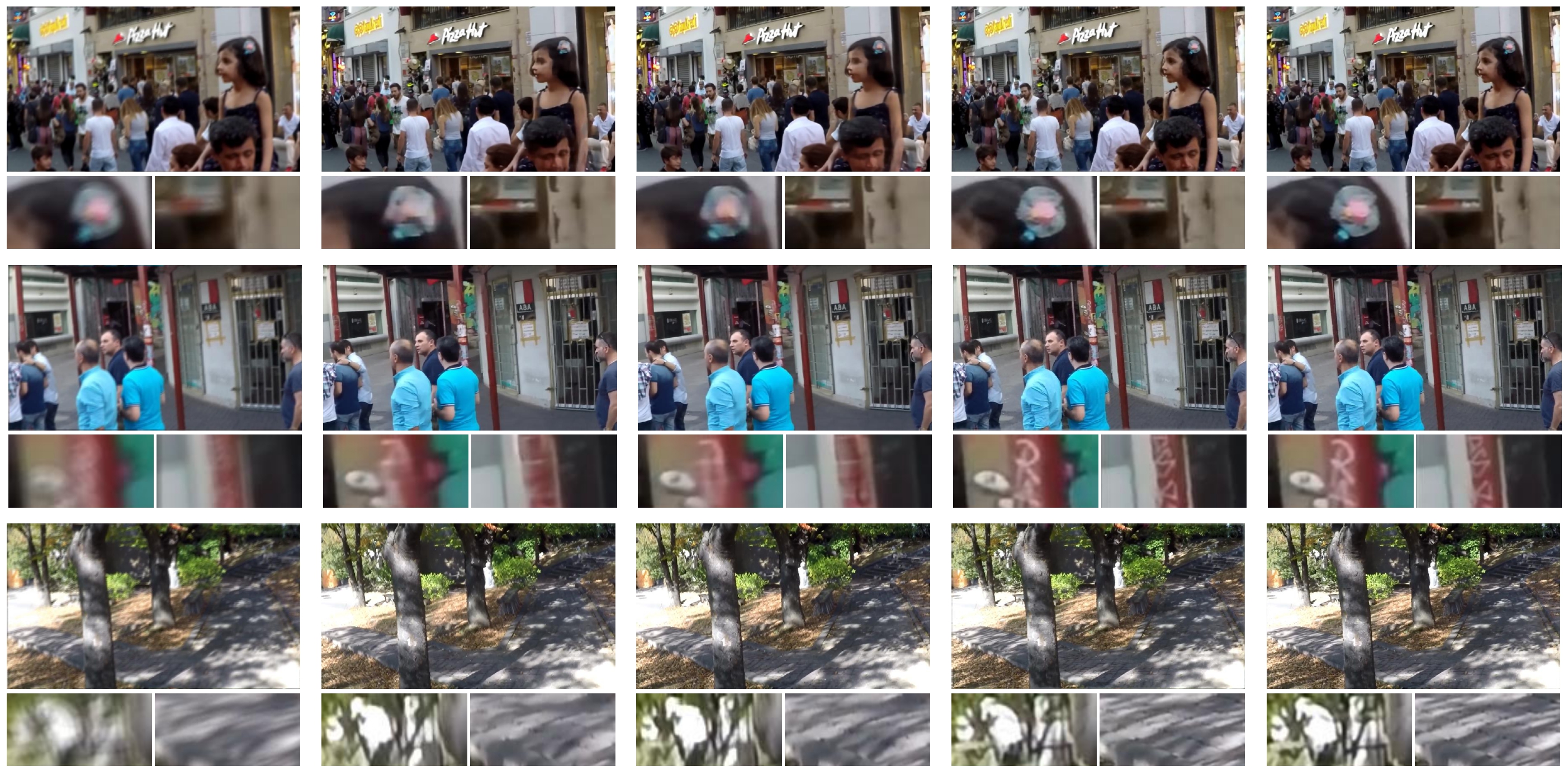}
  \caption{ Comparison among state-of-the-art non-blind and blind image deblurring methods. The ground-truth blur kernels are provided. From left to right: blurry images, results of  MSCNN \cite{nah2017deep}, DeblurGAN-v2 \cite{kupyn2019deblurgan}, DWDN \cite{dong2021deep} and USRNet \cite{Zhang_2020_CVPR}. MSCNN \cite{nah2017deep} and DeblurGAN-v2 \cite{kupyn2019deblurgan} are blind image deblurring methods. DWDN \cite{dong2021deep} and USRNet \cite{Zhang_2020_CVPR} are non-blind methods.}
  \label{fig_gopro_blind_non_blind}
\end{figure*}

\section{Performance Evaluation} 
\label{sec_experiments}

In this section, we discuss the performance evaluation of representative deep deblurring methods.

\begin{table*}[tb]
\tiny
  \centering 
    \caption{Performance evaluation of representative methods for video deblurring on two widely used datasets, DVD \cite{su2017deep} and REDS \cite{nah2019ntire}. MS stands for multi-scale.  In order to show the performance gap between video deblurring and single image deblurring methods, the table includes some single image deblurring methods.}
    \begin{tabular}{l | l l l l r  l }
    \toprule
    Dataset &  Method  & Framework & Layers \& block &  Loss & PSNR/SSIM & Characteristic \\
    \midrule
    \multirow{5}{*}{DVD \cite{su2017deep}} & Su~\etal~\cite{su2017deep} & U-Net & Convolution & $\mathcal{L}_{pix}$ & 30.05/0.920 & \textbf{Video}-based deep deblurring Net\\
     & Kim~\etal~\cite{hyun2017online} & Video, DAE & Recurrent, Residual & $\mathcal{L}_{pix}$ & 29.95/0.911 & \textbf{Video}, Dynamic temporal blending\\
     & Zhou~\etal~\cite{zhou2019spatio} & DAE & Resblock & $\mathcal{L}_{pix}$, $\mathcal{L}_{per}$ & 31.24/0.934 & \textbf{Video}, Filter Adaptive Net\\
     & Nah~\etal~\cite{nah2019recurrent} & DAE & Recurrent, Resblock & $\mathcal{L}_{pix}$,$\mathcal{L}_{regular}$ & 30.80/0.8991 & \textbf{Video}, RNN, Intra-frame iterations \\
     & Pan~\etal~\cite{pan2020cascaded} & Cascade, DAE & Resblock & $\mathcal{L}_{pix}$ & 32.13/0.9268 & \textbf{Video}, Sharpness Prior, optical flow \\
    \midrule
    \multirow{6}{*}{DVD \cite{su2017deep}} & Sun~\etal~\cite{sun2015learning} & DAE & Convolution & $\mathcal{L}_{pix}$ & 27.24/0.878 & CNN-based approach, MRF\\
     & Gong~\etal~\cite{gong2017motion} & DAE & Fully convolution & $\mathcal{L}_{pix}$, $\mathcal{L}_{flow}$ & 28.22/0.894 & Estimation of motion flow\\
     & Nah~\etal~\cite{nah2017deep} & MS, GAN & ResBlock & $\mathcal{L}_{per}$, $\mathcal{L}_{adv}$ & 29.51/0.912 & Multi-scale Net, adversarial loss\\
     & Kupyn~\etal~\cite{kupyn2018deblurgan} & Conditional GAN & ResBlock & $\mathcal{L}_{per}$, $\mathcal{L}_{adv}$ & 26.78/0.848 & Conditional GAN-based model\\
     & Zhang~\etal~\cite{zhang2018dynamic} & U-Net & Recurrent, Residual & $\mathcal{L}_{pix}$ & 30.05/0.922 & Spatially variant RNN \\
     & Tao~\etal~\cite{tao2018scale} & U-Net, MS & Recurrent, Dense & $\mathcal{L}_{pix}$ & 29.97/0.919 & Scale-recurrent Net \\
    \midrule
    \multirow{2}{*}{REDS \cite{nah2019ntire}}  & Su~\etal~\cite{su2017deep} & U-Net & Convolution & $\mathcal{L}_{pix}$ & 26.55/0.8066 & \textbf{Video}-based deep deblurring Net\\
     & Wang~\etal~\cite{wang2019edvr} & Video, DAE & Recurrent, ResBlock & $\mathcal{L}_{pix}$ & 34.80/0.9487 & \textbf{Video}, Deformable CNN\\
    \midrule
    \multirow{3}{*}{REDS \cite{nah2019ntire}} & Kupyn~\etal~\cite{kupyn2018deblurgan} & Conditional GAN & ResBlock & $\mathcal{L}_{per}$, $\mathcal{L}_{adv}$ & 24.09/0.7482 & Conditional GAN-based model\\
     & Nah~\etal~\cite{nah2017deep} & MS, GAN & ResBlock & $\mathcal{L}_{per}$, $\mathcal{L}_{adv}$ & 26.16/0.8249 & Multi-scale Net, adversarial loss\\
     & Tao~\etal~\cite{tao2018scale} & U-Net, MS & Recurrent, Dense & $\mathcal{L}_{pix}$ & 26.98/0.8141 & Scale-recurrent Net \\
    \bottomrule
    \end{tabular}%
    \label{table_compare_video}
\end{table*}%

\begin{figure*}[tb]
  \centering
  \tiny
\begin{tikzpicture}
\begin{axis}[
width=18cm,height=7cm, 
symbolic x coords={Su \cite{su2017deep}, Kim \cite{hyun2017online}, Zhou \cite{zhou2019spatio},Nah \cite{nah2019recurrent},Pan \cite{pan2020cascaded},  Sun \cite{sun2015learning}, Gong \cite{gong2017motion}, Nah \cite{nah2017deep}, Kupyn \cite{kupyn2018deblurgan}, Zhang \cite{zhang2018dynamic}, Tao \cite{tao2018scale}},
xtick=data
]
\addplot[ybar,fill=blue] coordinates { 
    (Su \cite{su2017deep}, 30.05)
    (Kim \cite{hyun2017online}, 29.95)
    (Zhou \cite{zhou2019spatio}, 31.24)
    (Nah \cite{nah2019recurrent}, 30.80)
    (Pan \cite{pan2020cascaded},32.13)
    (Sun \cite{sun2015learning},   27.24)
    (Gong \cite{gong2017motion},  28.22)
    (Nah \cite{nah2017deep},   29.51)
    (Kupyn \cite{kupyn2018deblurgan}, 26.78)
    (Zhang \cite{zhang2018dynamic}, 30.05)
    (Tao \cite{tao2018scale}, 29.97)
};
\end{axis}
\end{tikzpicture}
\caption{{\bf Comparison among state-of-the-art video deblurring methods in terms of PSNR on the DVD dataset.} } 
\label{bar_graph_dvd}
\end{figure*}
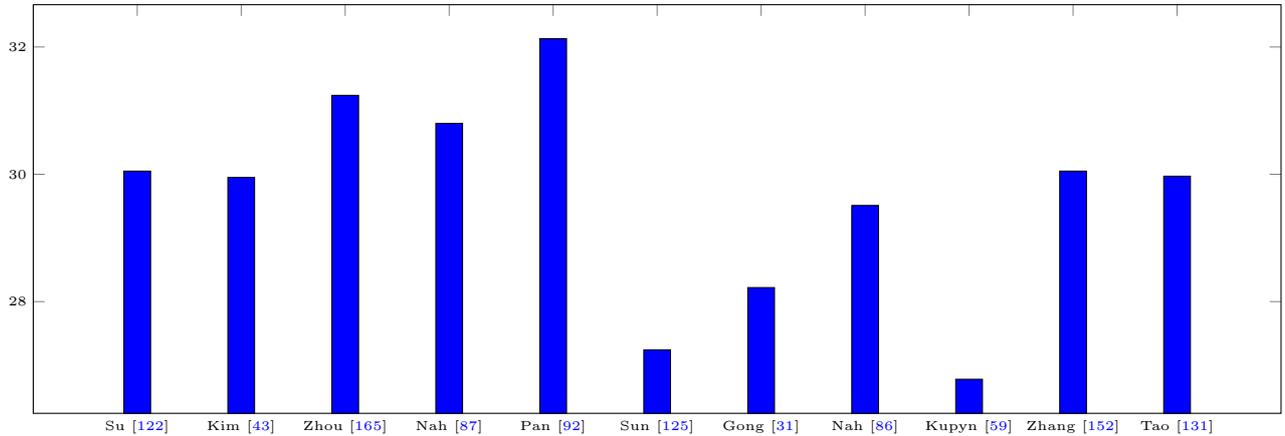

\begin{table}[tb]
  \centering 
    \caption{Speed of representative methods for deep video deblurring. The numbers are quoted from \cite{nah2019recurrent}.}
    \begin{tabular}{l | c}
    \toprule
    Method & Speed (fps) \\
    \hline
    Su~\etal~\cite{su2017deep} & 1.72 \\
    Kim~\etal~\cite{hyun2017online}  & 9.24 \\
    Nah~\etal~\cite{nah2019recurrent}  & 28.8 \\
     \bottomrule
    \end{tabular}%
    \label{table_compare_video_speed}
\end{table}%

\subsection{Single Image Deblurring}

We summarize representative methods in Table~\ref{table_compare_image}, which compares the performance on three popular single image deblurring datasets, the GoPro \cite{nah2017deep} dataset, and the datasets from K\" ohler~\etal~\cite{kohler2012recording} and Shen~\etal~\cite{shen2019human}. Note that all results are obtained from the respective papers.
For the results of single image deblurring on the GoPro dataset and the results of video deblurring on the DVD dataset, we additionally provide bar graphs for easier comparison, as shown in Fig.~\ref{bar_graph_gopro} and Fig.~\ref{bar_graph_dvd}.

The methods developed by Sun~\etal~\cite{sun2015learning}, Gong~\etal~\cite{gong2017motion} and Zhang~\etal~\cite{zhang2018dynamic} are three early deep deblurring networks based on CNN and RNN, showing that deep learning based methods achieve competitive results.
Using coarse-to-fine schemes, Nah~\etal~\cite{nah2017deep}, Tao~\etal~\cite{tao2018scale}, and Gao~\etal~\cite{gao2019dynamic} designed  multi-scale networks and achieved better performance compared to single-scale deblurring networks as the coarse deblurring networks provide a better prior for higher resolutions.
In addition to the multi-scale strategy, GANs have been employed to produce more realistic deblurred images \cite{kupyn2018deblurgan,kupyn2019deblurgan,zhang2020deblurring}. However,  GAN-based models achieve poorer results in terms of PSNR and SSIM metrics on the GoPro dataset \cite{nah2017deep} as shown in Table~\ref{table_compare_image}.
Purohit~\etal~\cite{purohit2019region} applied an attention mechanism to the  deblurring network and achieved state-of-the-art performance on the GoPro dataset. The results on Shen~\etal's dataset \cite{shen2019human}  demonstrate the effectiveness. 
Networks using ResBlocks \cite{nah2017deep} and DenseBlocks \cite{gao2019dynamic,zhang2020deblurring} also achieve better performance than previous networks which directly stack convolutional layers \cite{gong2017motion,sun2015learning}. 
For non-UHD motion deblurring, the multi-patch architecture has the advantage of restoring images better than the multi-scale and GAN-based architectures. 
%
%
For other architectures, such as multi-scale networks, the small-scale version of images provides less details, and thus these methods cannot improve the performance significantly.

Fig. \ref{ablation_structure} shows sample results of single image deblurring  from the GoPro dataset. 
We compare two multi-scale methods \cite{nah2017deep,tao2018scale} and two GAN-based methods \cite{kupyn2019deblurgan,zhang2020deblurring} in the experiments. 
Despite the differences in model architectures, all these methods perform well on this dataset. 
Meanwhile, methods using similar architectures can yield different results, \eg,  Tao \textit{et al.} and Nah \textit{et al.} \cite{nah2017deep}, which are based on multi-scale architectures. Tao \textit{et al.} use recurrent operations between different scales and achieve sharper deblurring results.

We also analyze the performance on the GoPro dataset using the LPIPS metric. 
Results of representative methods are shown in Table~\ref{table_compare_image_lpips}. Under this metric, Tao \textit{et al.} \cite{tao2018scale} performs worse than Nah \textit{et al.} \cite{nah2017deep} and Kupyn \textit{et al.} \cite{kupyn2018deblurgan}. 
This is because the LPIPS measures the perceptual similarity rather than the pixel-wise similarity measures like L1.
%
Specifically, Tao \textit{et al.} \cite{tao2018scale} outperforms the Nah \textit{et al.} \cite{nah2017deep} and Kupyn \textit{et al.} \cite{kupyn2018deblurgan} in terms of PSNR and SSIM on the GoPro dataset.
The results show that we may draw different conclusions when various metrics are used. 
Therefore, it is critical to evaluate deblurring methods using different quantitative metrics. 
As evaluation metrics optimize different criteria, the design choice is task-dependent.
There is a perception {\it vs.} distortion tradeoff in image reconstruction tasks, and it has been shown that these two measures are at odds with each other, see~\cite{blau2018tradeoff}.
To evaluate methods in fairly, one may design a combined cost function including measures such as PSNR, SSIM, FID, LPIPS, NIQE, and carry out user studies.

To analyze the effectiveness of different loss functions, we design a common backbone of several ResBlocks (provided by DeblurGAN-v2). 
Different loss functions, including L1, L2, perceptual loss, GAN loss, and RaGAN loss \cite{zhang2020deblurring}, and their various combinations are used for training.
Similar to the settings in DeblurGAN-v2 \cite{kupyn2019deblurgan}, we train each model for $200$ epochs on the GoPro dataset.
The learning rate scheduler corresponds to that in DeblurGAN-v2. 
Experimental results are reported in Table~\ref{table_ablation_loss_fuction}. 
In general, the evaluated models perform better when combining reconstruction and perceptual loss functions. 
However, using GAN-based loss functions does not necessarily improve PSNR or SSIM, which is consistent with the findings in prior reviews \cite{ledig2017photo}. 
In addition, the same model using GAN loss or RaGAN loss achieves similar deblurring performance.

In order to evaluate the performance difference between non-blind and blind single image deblurring methods, we conduct a study comparing representative non-blind (FDN \cite{kruse2017learning} and RGDN \cite{gong2020learning}) and blind single image deblurring methods \cite{gao2019dynamic,kupyn2019deblurgan,nah2017deep,tao2018scale,zhang2020deblurring} on the RWBI dataset. Considering that the RWBI dataset does not provide the ground-truth blur kernels, we use \cite{xu2013unnatural} to estimate blur kernels for non-blind image deblurring. Results are presented in Table~\ref{table_compare_image_niqe_brisque} using the NIQUE and BRIQUE metrics. Overall, the blind deblurring methods outperform the non-blind methods since non-blind approaches require explicitly estimating the blur kernel. In practice, estimating the blur kernel is still a challenging task. If a kernel is not well estimated, it will negatively affect the image restauration task. Fig.~\ref{fig_rwbi_blind_non_blind} shows sample results from the RWBI dataset.

To analyze the performance gap between blind and non-blind deblurring methods, we synthesize a new dataset with ground-truth blur kernels. 
Specifically, we use the blur kernels applied in \cite{Zhang_2020_CVPR} (including 4 isotropic Gaussian kernels, 4 anisotropic Gaussian kernels from \cite{zhang2018learning}, and 4 motion blur kernels from \cite{boracchi2012modeling,levin2009understanding}) to generate blurry images based on the sharp images from the GoPro dataset \cite{nah2017deep}. The generation method is based on Eq. \ref{f_normal} using Gaussian noise.
The blur kernels and blurry images are the input to non-blind image deblurring methods (DWDN \cite{dong2021deep} and USRNet \cite{Zhang_2020_CVPR}), while the input to blind image deblurring methods (MSCNN \cite{nah2017deep} and SRN \cite{tao2018scale}) are only blurry images. 
Table \ref{table_compare_image_non_blind} and Fig. \ref{fig_gopro_blind_non_blind} show quantitative and qualitative results, respectively. 
Overall, non-blind image deblurring methods can achieve better performance than blind image deblurring approaches if the ground-truth blur kernels are provided.

\begin{table}
  \centering 
    \caption{Performance evaluation of different training sets \cite{martin2001database,nah2017deep,rim2020real}. The image deblurring model is Tao \etal \cite{tao2018scale}, and the test set is from RealBlur-J dataset \cite{rim2020real}. The values are reported in \cite{rim2020real}.}
    \begin{tabular}{l | c c}
    \toprule
    Training set & PSNR & SSIM \\
    \hline
    RealBlur-J   & 31.02 & 0.8987 \\
    GoPro  & 28.56 & 0.8674 \\
    BSD-B   & 28.68 & 0.8675 \\
    RealBlur-J + GoPro  & 31.21 & 0.9018 \\
    RealBlur-J + BSD-B  & 31.30 & 0.9058 \\
    RealBlur-J + BSD-B + GoPro  & 31.37 & 0.9063 \\
     \bottomrule
    \end{tabular}%
    \label{table_ablation_data}
\end{table}%

For a better understanding of the role of different training datasets, 
we train the SRN model \cite{tao2018scale} on three different public datasets (RealBlur-J \cite{rim2020real}, GoPro \cite{nah2017deep}, and BSD-B \cite{martin2001database,rim2020real}) separately. 
In addition, we also train the SRN model on the combinations of ``RealBlur-J + GoPro'', ``RealBlur-J + BSD-B'', and ``RealBlur-J + BSD-B + GoPro''. 
Results are reported in Table \ref{table_ablation_data}. 
The models using diverse data perform better than those only using one type of training data.
In particular, the model using the combination of three datasets achieves the highest PSNR and SSIM values on the RealBlur-J testing dataset.

Considering that modern mobile devices allow capturing ultra-high-definition (UHD) images, we synthesize a new dataset to study the performance of different architectures on UHD image deblurring. 
We use a Sony RX10 camera to capture 500 and 100 sharp images with 4K resolution as training and testing sets, respectively. 
Next, we use 3D camera trajectories to generate blur kernels and synthesize corresponding blurry images by convolving sharp images with large blur kernels  (sizes $111 \times 111$, $131 \times 131$, $151 \times 151$, $171 \times 171$, $191 \times 191$). 
We use two multi-scale networks (MSCNN, SRN), two GAN-based networks (DeblurGAN, DeblurGAN-v2) and one multi-patch networks (DMPHN) for experiments.
Table \ref{table:results_conv} shows the results of these representative deblurring methods. 
The results show that UHD image deblurring is a more challenging task and multi-scale architectures achieve better performance in terms of PSNR and SSIM.
For UHD motion deblurring, multi-scale architectures achieve better performance. 
As UHD images have higher resolution, downsampled versions at $1/4$ scale still contain sufficient detail. 
Multi-scale architectures take UHD blurry images at $1/4$ of the original resolution as input and generate the corresponding sharp versions. During the training stage, the images with $1/4$ resolution can provide additional information for training and thus improve the performance of deblurring networks.

\begin{table}[tb]
  \centering 
    \caption{Performance evaluation of representative methods for UHD image deblurring.}
    \begin{tabular}{c | c c }
    \hline
    Method &  PSNR  & SSIM \\
    \hline
    DeepDeblur  \cite{nah2017deep} & 21.12 & 0.6226 \\
    DeblurGAN \cite{kupyn2018deblurgan} & 19.25 & 0.5477 \\
    SRN \cite{tao2018scale} & 21.25 & 0.6233  \\
    DeblurGAN-v2 \cite{kupyn2019deblurgan} & 19.99 & 0.5865 \\
    DMPHN \cite{zhang2019deep2} & 20.98 & 0.6217 \\
    \bottomrule
    \end{tabular}%
    \label{table:results_conv}
\end{table}%

\begin{table}[tb]
  \centering 
    \caption{Performance evaluation of representive methods for defocus deblurring.}
    \begin{tabular}{c | c c }
    \hline
    Method &  PSNR  & SSIM \\
    \hline
    DeepDeblur  \cite{nah2017deep} & 19.78 & 0.7107   \\
    DeblurGAN \cite{kupyn2018deblurgan} & 19.12 & 0.6263 \\
    SRN \cite{tao2018scale} & 20.45 & 0.7557 \\
    DeblurGAN-v2 \cite{kupyn2019deblurgan} & 20.02 & 0.6896 \\
    DMPHN \cite{zhang2019deep2} & 20.12 & 0.7467 \\
    \bottomrule
    \end{tabular}%
    \label{table:results_defocus}
\end{table}%

In order to analyze the performance of different architectures on defocus deblurring, we create another dataset and conduct numerous experiments. 
Specifically, we use a Sony RX10 camera to capture 500 pairs of sharp and blurry images for training, and 100 pairs of sharp and blurry images for testing. 
The image resolution is $900 \times 600$ pixels. 
Similarly, two multi-scale networks (MSCNN, SRN), two GAN-based networks (DeblurGAN, DeblurGAN-v2) and one multi-patch network (DMPHN) are evaluated on this dataset. 
Table \ref{table:results_defocus} shows the results of the above methods. 
Compared with averaging-based motion deblurring, GAN-based architectures can achieve better performance on defocus deblurring. 
Multi-scale and multi-patch architectures do not show significant improvements in terms of PSNR and SSIM.
For defocus deblurring, results show that these two networks (with and without GAN framework) do not show significant differences in terms of PSNR and SSIM. 
However, for motion deblurring, GAN-based architectures yield lower PSNR and SSIM values. 
This may be attributed to the fact that GAN-based architectures paying attention to whole images using the adversarial loss function. In comparison, methods without GANs consider the pixel-level error  (L1, L2), and thus ignore the whole image.
We provide a run-time comparison of representative methods in Table~\ref{table_compare_image_speed}.

\subsection{Video Deblurring}

In this section, we compare recent video deblurring approaches on two widely used video deblurring datasets, the DVD \cite{su2017deep} and REDS \cite{nah2019ntire} datasets, see Table~\ref{table_compare_video}. 
Deep auto-encoders~\cite{su2017deep} are the most commonly used architecture for deep video deblurring methods. 
Similar to single image deblurring methods, the convolutional layer and ResBlocks \cite{he2016deep} are the most important components.
The recurrent structure \cite{kim2018spatio} is used to extract temporal information from neighboring blurry images, which is the main difference to single image deblurring networks.

Pixel-wise loss functions are employed in most video deblurring methods, but, similar to single image deblurring, perceptual loss \cite{zhou2019davanet} and adversarial loss \cite{zhang2018adversarial} have also been used. 

While single image deblurring methods can be applied to videos, they are outperformed by video-based methods that exploit temporal information.
Su~\etal~\cite{su2017deep} develop a blurry video dataset (DVD) and introduce a CNN-based video deblurring network, which outperforms non-deep learning based video deblurring approaches.
By using intra-frame iterations, the RNN-based video deblurring network by Nah~\etal~\cite{nah2019ntire} achieves better performance than Su~\etal~\cite{su2017deep}. 
Zhou~\etal~\cite{zhou2019spatio} proposed the STFAN module, which better incorporates information from preceding frames into the deblurring process of the current frame. In addition, a filter adaptive convolutional (FAC) layer is employed for aligning the deblurred features from these frames. Pan~\etal~\cite{pan2020cascaded} achieve state-of-the-art performance on the DVD dataset by using a cascaded network and PWC-Net to estimate optical flow for calculating sharpness priors \cite{sun2018pwc}. A DAE network uses the priors and blurry images to estimate sharp images. Run times of representative methods are shown in Table~\ref{table_compare_video_speed}.

\section{Domain-specific Deblurring}
\label{sec_domain}

While most deep learning based methods are designed for deblurring {\it generic} images, \ie~common natural and man-made scenes, some methods have studied the deblurring problem in specific domains or settings, \eg, faces and texts.

\subsection{Face Image Deblurring}

\begin{figure}[!tb]
  \centering
    \includegraphics[width=0.9\linewidth ]{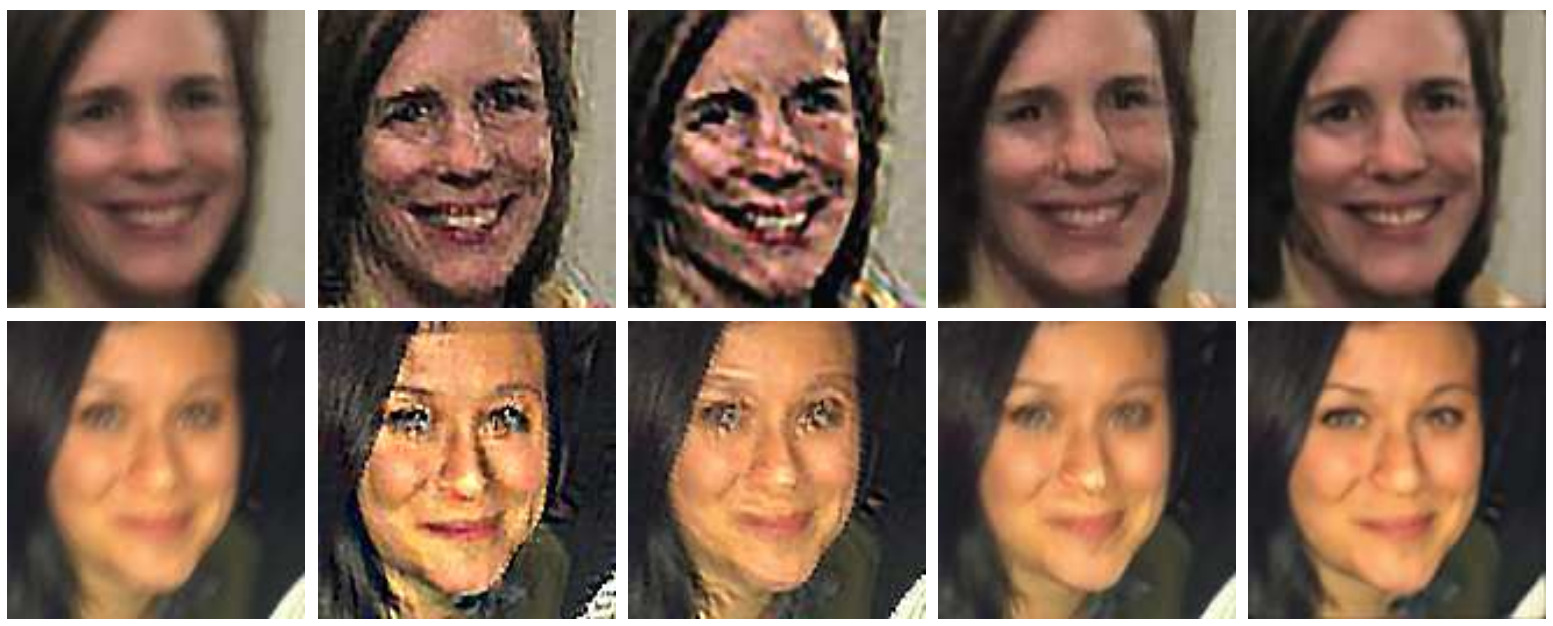}
\caption{Face deblurring examples. From left to right: input images, results obtained by the methods in \cite{xu2013unnatural}, \cite{pan2014deblurring_face}, \cite{nah2017deep} and \cite{shen2018deep}, respectively.}
  \label{application_face}
\end{figure}

Early face image deblurring methods used key structures of facial images \cite{hacohen2013deblurring,pan2014deblurring_face}. Recently, deep learning solutions have dominated the development of face deblurring by exploiting specific facial characteristics \cite{jin2018learning,ren2019face,shen2018deep,xu2017learning}. Shen~\etal~\cite{shen2018deep} use semantic information to guide the process of face deblurring. Pixel-wise semantic labels are extracted via a parsing network and  serve as priors to the deblurring network to restore sharp faces. Jin~\etal~\cite{jin2018learning} develop an end-to-end network with a resampling convolution operation to widen the receptive field. Chrysos~\etal~\cite{chrysos2019motion} propose a two-step architecture, which first restores the low frequencies and then restores the high frequencies to ensure the outputs lie on the natural image manifold. To address the task of face video deblurring, Ren~\etal~\cite{ren2019face}  explore 3D facial priors. A deep 3D face reconstruction network generates a textured 3D face for the blurry input, and a face deblurring branch recovers the sharp face under the guidance of the posed-aligned face. Figure~\ref{application_face} shows deblurring results from two non-deep methods \cite{pan2014deblurring_face,xu2013unnatural} and two deep learning based methods \cite{nah2017deep,shen2018deep}.

\subsection{Text Image Deblurring}

\begin{figure}[!tb]
  \centering
    \includegraphics[width=1\linewidth ]{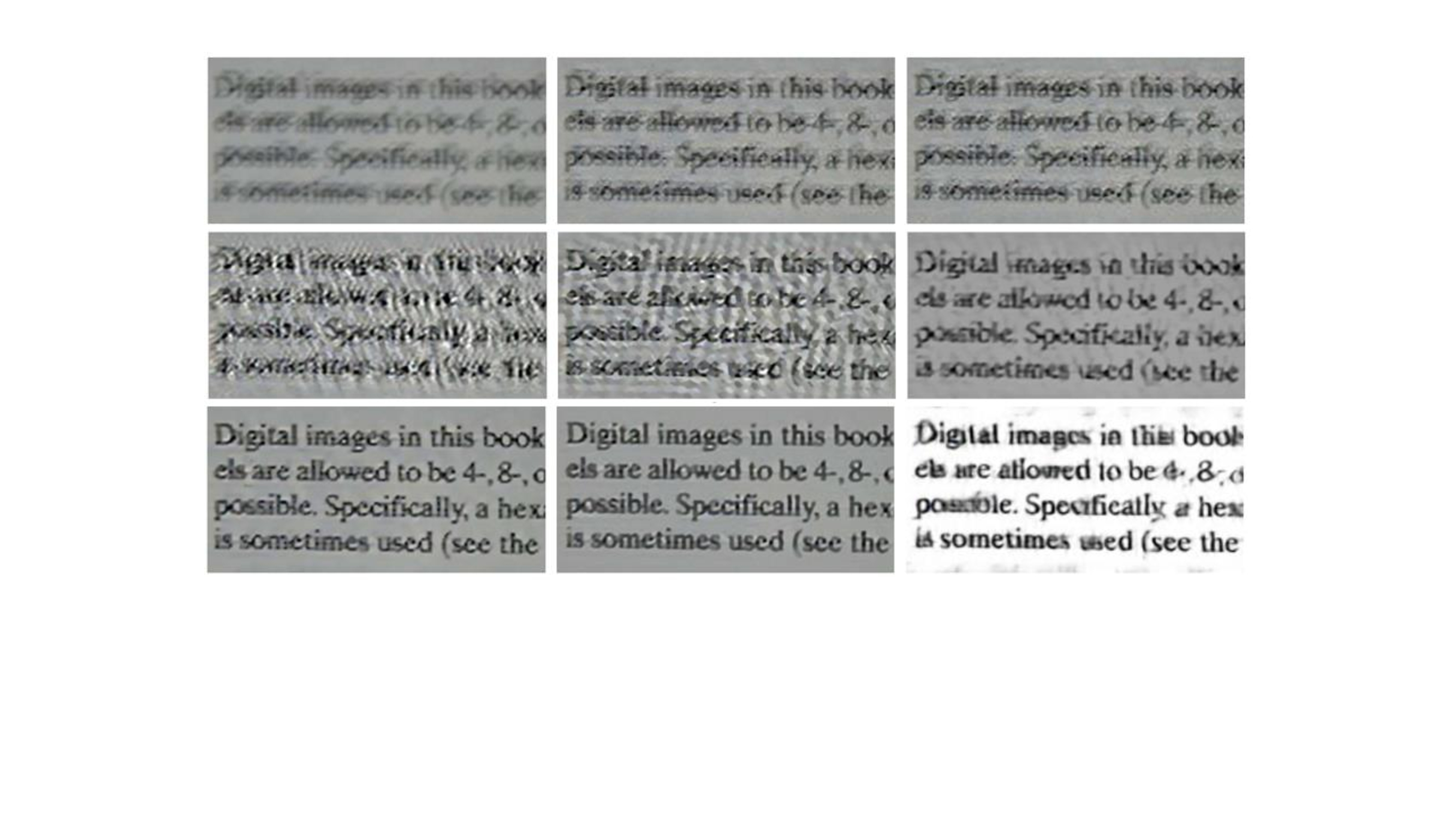}
\caption{Text deblurring examples. (From left to right, top to bottom) input image,  results from \cite{xu2010two} \cite{xu2013unnatural}, \cite{cho2009fast}, \cite{zhong2013handling}, \cite{chen2011effective}, \cite{cho2012text}, \cite{pan2014deblurring} and \cite{hradivs2015convolutional}, respectively.}
  \label{application_text}
\end{figure}

Blurred text images impact the performance of optical character recognition (OCR), \eg, when reading documents, displays, or street signs.
Generic image deblurring methods are not well suited for text images. 
In early work Panci~\etal~\cite{panci2003multichannel} model the text image as a random field to remove blur via blind deconvolution algorithms~\cite{fiori1999blind}. Pan~\etal~\cite{pan2014deblurring} deblur text images via an $L_0$-regularized prior based on intensity and gradient. More recently, deep learning methods, \eg~the method by Hradi{\v{s}}~\etal~\cite{hradivs2015convolutional} have shown to effectively remove out-of-focus and motion blur. Xu~\etal~\cite{xu2017learning} adopt a GAN-based model to learn a category-specific prior for the task, designing a multi-class GAN model and a novel loss function for both face and text images. Figure~\ref{application_text} shows the deblurring results from several non-deep methods \cite{chen2011effective,cho2012text,cho2009fast,pan2014deblurring,xu2010two,xu2013unnatural,zhong2013handling} and the deep learning based method in \cite{hradivs2015convolutional}, showing that \cite{hradivs2015convolutional} achieves better deblurring results on text images.

\begin{figure}[!tb]
  \centering
    \includegraphics[width=1\linewidth ]{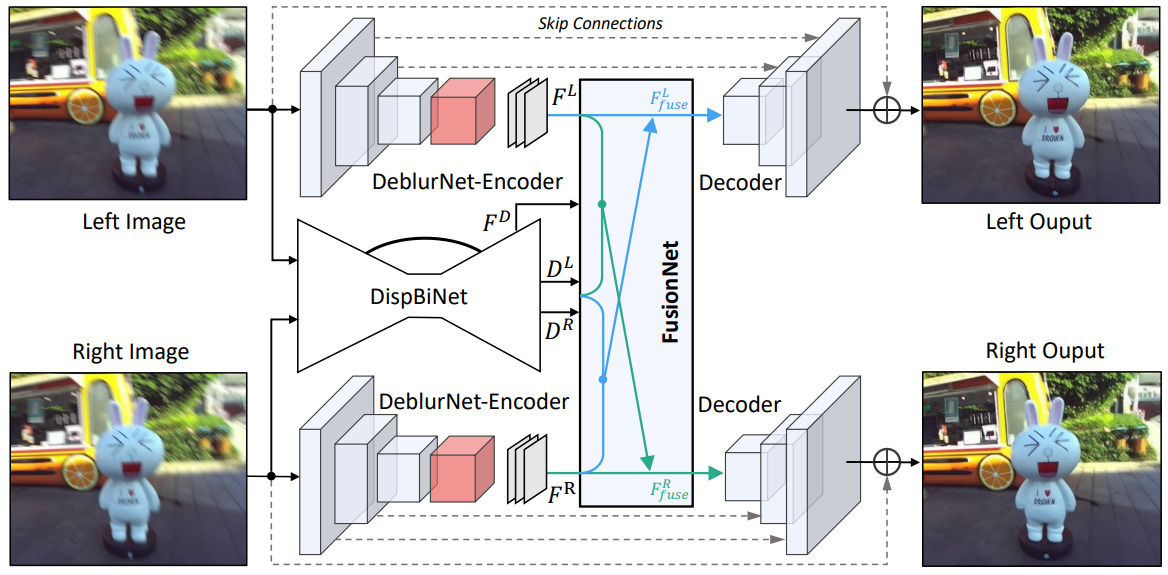}
\caption{Stereo image deblurring example \cite{zhou2019davanet}. Two stereo blurry images are fed into deep deblurring networks to generate their corresponding sharp images.}
  \label{application_stereo}
\end{figure}

\subsection{Stereo Image Deblurring}

Stereo vision has been widely used to achieve depth perception \cite{godard2017unsupervised} and 3D scene understanding \cite{eslami2016attend}. When mounting a stereo camera on a moving platform, vibration will lead to blur, negatively affecting the subsequent stereo computation \cite{sellent2016stereo}.
To alleviate this issue, Zhou~\etal~\cite{zhou2019davanet} proposed a deep stereo deblurring method to make use of depth information, and varying information in corresponding pixels across two stereo images. 

Depth information helps estimate the spatially-varying blur as it provides prior knowledge that nearer points are more blurry than farther points, in the case of translational motion, which is a common real-world scenario. Further, the blur of corresponding pixels in two stereo images is different, which allows using the sharper pixel in the final deblurred images.
In addition to the above tasks, other domain-specific deblurring tasks include extracting a video sequence from a single blurred image \cite{jin2018learning,purohit2019bringing}, synthesizing high-frame-rate sharp frames \cite{shen2020blurry}, and joint deblurring and super-resolution \cite{zhang2018gated}.

\section{Challenges and Opportunities}
\label{sec_challenge}

Despite the significant progress of image deblurring algorithms on benchmark datasets, recovering clear images from a real-world blurry input remains challenging \cite{nah2017deep,su2017deep}.
In this section, we summarize key limitations and discuss  possible approaches and research opportunities.

{\flushleft \textbf{Real-world data.}} 
There are three main reasons that deblurring algorithms do not perform well on real-world images.

First, most deep learning based methods require paired blurry and sharp images for training, where the blurry inputs are artificially synthesized. However, there is still a gap between these synthetic images and real-world blurry images as the blur models (\eg, Eq. \ref{f_normal} and \ref{blurr_process2}) are oversimplified \cite{brooks2019learning,chen2018reblur2deblur,zhang2020deblurring}.
Models trained on synthetic blur achieve excellent performance on synthetic test samples, but tend to perform worse on real-world images. 
One feasible approach to obtain better training samples is to build better imaging systems \cite{rim2020real}, \eg, by capturing the scenes using different exposure times.
Another option is to develop more realistic blur models that can synthesize more realistic training data.

Second, real-world images are not only corrupted due to blur artifacts, but also due to quantization, sensor noise, and other factors like low-resolution \cite{shen2018deep,zhang2018gated}. One way to address this problem is to develop a unified image restoration model to recover high-quality images from  the inputs corrupted by various nuisance factors.

Third, deblurring models trained on general images may perform poorly on images from domains that have different characteristics. 
Specifically, it is challenging for general methods to recover sharp images of faces or text while maintaining the identity of a particular person or characters in the text, respectively.

{\flushleft \textbf{Loss functions.}} While numerous loss functions have been developed in the literature, it is not clear how to use the right formulation for a specific scenario.
For example, as Table \ref{table_ablation_loss_fuction} shows that an image deblurring model trained with L1 loss may achieve better performance than models trained with L2 loss on the GoPro dataset.
However, the same network trained with L1 and perceptual loss functions is worse than the models trained with L2 and perceptual loss functions. 

{\flushleft \textbf{Evaluation metrics.}} 
The most widely used evaluation metrics for image deblurring are PSNR and SSIM. However, the PSNR metric is closely related to the MSE loss, which favors over-smoothed predictions. Therefore, these metrics cannot accurately reflect perceptual quality. Images with lower PSNR and SSIM values can have better visual quality \cite{ledig2017photo}. The Mean Opinion Score (MOS) is an effective measure of perceptual quality. However, this metric is not universal and cannot be easily reproduced as it requires a user study. Therefore, it is still challenging to derive evaluation metrics that are consistent with the human visual response.

{\flushleft \textbf{Data and models.}} Both data and models play important roles in obtaining favorable deblurring results. 
In the training stage, high-quality data is important to construct a strong image deblurring model.  
However, it is difficult to collect large-scale high-quality datasets with ground-truth.
Usually, it requires using to use two cameras (with proper configurations) to capture real pairs of training samples, and thus the amount of these high-quality data is small with small diversity. 
While it is easier to generate a dataset with synthesized images, deblurring models developed based on such data usually perform worse than those built upon high-quality real-world samples.
Therefore, constructing a large number of high-quality datasets is an important and challenging task.
In addition, high-quality datasets should contain diverse scenarios, in terms of types of objects, motion, scenes, and image resolution. 
Deblurring network models have been mainly designed based on empirical knowledge. 
Recent neural architecture search methods, such as AutoML \cite{liu2019darts,pmlr-v80-pham18a,zoph2017neural}, may also be applied to the deblurring task. 
In addition, transformers \cite{vaswani2017attention} have achieved great success in various computer vision tasks. How to design more powerful backbones using Transformer may be an opportunity.

{\flushleft \textbf{Computational cost.}} Since many current mobile devices support capturing 4K UHD images and videos, we test several state-of-the-art deep learning based deblurring methods. However, we found that most these deep learning based deblurring methods cannot handle 4K resolution images at high speed. For example, on a single NVIDIA Tesla V100 GPU, Tao \textit{et al.} \cite{tao2018scale}, Nah \textit{et al.} \cite{nah2017deep} and Zhang \textit{et al.} \cite{zhang2020deblurring} 
take approximately 26.76, 28.41, 31.62 seconds, respectively, to generate a 4K deblurred image. Therefore, efficiently restoring high-quality UHD images remains an open research topic.
Note that most existing deblurring networks are evaluated on desktops or servers equipped with high-end GPUs. 
However, it requires significant effort to  develop efficient deblurring algorithms directly on mobile devices. One recent example is the work by Kupyn~\etal~\cite{kupyn2019deblurgan}, which proposes a MobileNet-based model for more efficient deblurring.

{\flushleft  \textbf{Unpaired learning.}} Current deep deblurring methods rely on pairs of sharp images and their blurry counterparts. However, synthetically blurred images do not represent the range of real-world blur. 
To make use of unpaired example images,
Lu~\etal~\cite{lu2019unsupervised} and Madam~\etal~\cite{madam2018unsupervised} recently proposed two unsupervised domain-specific deblurring models. 
Further improving semi-supervised or unsupervised methods to learn deblurring models appears to be a promising research direction.

\section*{Acknowledgment}
This research was funded in part by the NSF CAREER
Grant \#1149783, ARC-Discovery grant  projects (DP 190102261 and DP220100800), and a Ford Alliance URP grant.

{\small
\bibliographystyle{spmpsci}
\bibliography{egbib}
}

\end{document}